\documentclass[letterpaper]{article} 
\usepackage{aaai24}  
\usepackage{times}  
\usepackage{helvet}  
\usepackage{courier}  
\usepackage[hyphens]{url}  
\usepackage{graphicx} 
\usepackage{natbib}  
\usepackage{caption} 
\usepackage{algorithm}
\usepackage{algorithmic}

\usepackage{newfloat}
\usepackage{listings}
\DeclareCaptionStyle{ruled}{labelfont=normalfont,labelsep=colon,strut=off} 
\floatstyle{ruled}
\newfloat{listing}{tb}{lst}{}
\floatname{listing}{Listing}
\usepackage{xcolor,colortbl}
\usepackage{subcaption}
\usepackage{tabularray}

\usepackage{tabularx}
\definecolor{oursrow}{HTML}{E6F2F4}

\definecolor{green}{HTML}{008015}
\definecolor{lightgray}{rgb}{0.9, 0.9, 0.9}

\definecolor{darker}{HTML}{fdae79}
\definecolor{lighter}{HTML}{fedfca}

\definecolor{g1}{HTML}{EECDCD}
\definecolor{g2}{HTML}{F8E6D0}
\definecolor{g3}{HTML}{FDF3D0}
\definecolor{g4}{HTML}{DCE9D5}
\definecolor{g5}{HTML}{D3E1F1}
\usepackage{pifont}
\newcommand{\cmark}{\ding{51}}%

\usepackage{siunitx}
\usepackage[english]{babel}

\usepackage{array, multirow, bigdelim, makecell, booktabs} 
\usepackage[framemethod=TikZ]{mdframed}
\usepackage{enumitem}
\usepackage{amsmath,amssymb,stmaryrd}   

\usepackage{amsthm}

\theoremstyle{definition}
\newtheorem{definition}{Definition}[section]

\theoremstyle{remark}

\title{Stable Diffusion Exposed: Gender Bias from Prompt to Image}
\author {
    Yankun Wu, 
    Yuta Nakashima, 
    Noa Garcia 
}
\affiliations {
    Osaka University\\
    \tt\{yankun@is., n-yuta@, noagarcia@\}ids.osaka-u.ac.jp
}

\begin{document}

\maketitle

\begin{figure*}[t]
\centering
  \includegraphics[clip, width=0.98\textwidth]{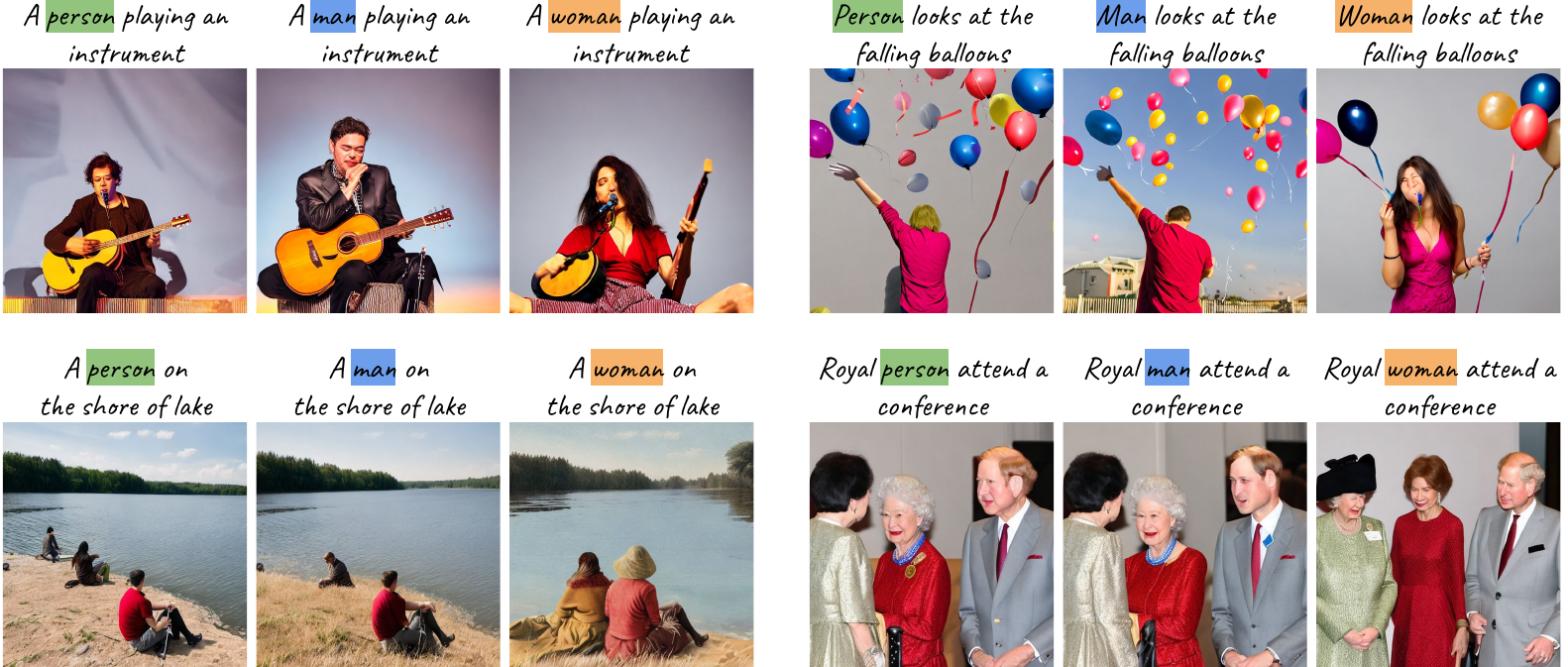}
    \captionsetup{type=figure}
    \vspace{-2pt}
  \captionof{figure}{
  We use free-form triplet prompts to analyze the influence of gender indicators on the overall image generation process. We show that 1) gender indicators influence the generation of objects (left) and their layouts (right), and 2) the use of gender \textit{neutral} words tends to produce images more similar to those prompted by \textit{masculine} indicators rather than \textit{feminine} ones.
  }
  \label{fig:first}
\end{figure*}%

\begin{abstract}
Several studies have raised awareness about social biases in image generative models, demonstrating their predisposition towards stereotypes and imbalances. This paper contributes to this growing body of research by introducing an evaluation protocol that analyzes the impact of gender indicators at every step of the generation process on Stable Diffusion images. Leveraging insights from prior work, we explore how gender indicators not only affect gender presentation but also the representation of objects and layouts within the generated images. Our findings include the existence of differences in the depiction of objects, such as instruments tailored for specific genders, and shifts in overall layouts. We also reveal that neutral prompts tend to produce images more aligned with masculine prompts than their feminine counterparts. We further explore where bias originates through representational disparities and how it manifests in the images via prompt-image dependencies, and provide recommendations for developers and users to mitigate potential bias in image generation. 
\end{abstract}    
\section{Introduction}
\label{sec:intro}

Text-to-image generation models have gained significant attention due to their remarkable generative capabilities. 
Cutting-edge models, such as Stable Diffusion \cite{rombach2022high} and DALL-E 2 \cite{ramesh2022hierarchical}, have demonstrated outstanding success in generating high-fidelity images based on natural language inputs. However, due to their widespread applications across different domains and their easy accessibility, concerns about the social impact of data \cite{birhane2021multimodal,garcia2023uncurated,birhane2023into}, bias \cite{bianchi2023easily,luccioni2023stable,ungless2023stereotypes}, privacy \cite{carlini2023extracting, katirai2023situating}, or intellectual property \cite{somepalli2023diffusion,wang2023evaluating} have surfaced. 
This work focuses on the automatic evaluation of gender bias in Stable Diffusion models. 

Previous studies have shown that certain adjectives \cite{luccioni2023stable} or professions \cite{luccioni2023stable} can lead to the generation of stereotypes regarding the demographic attributes of faces. 
However, beyond the regions depicting faces, the remaining areas of the generated image may also exhibit disparities between different genders \cite{bianchi2023easily}. Figure \ref{fig:first} shows triplets of generated images with prompts differing by only one word in the gender indicators.\footnote{Gender indicators refer to words that indicate the gender of a person.} The representation of the person in the images adapts accordingly, but the context surrounding the individual (e.g., different musical instruments on the left) and the layout of the image (e.g., on the right) also undergo alterations, even when these changes are not explicitly mentioned in the prompt. This reveals that gender bias does not only manifest within areas depicting people but is sustained in the broader context of the entire image. 
Thus, while demographic bias in text-to-image generation models has been consistently reported \cite{ungless2023stereotypes, bianchi2023easily, garcia2023uncurated, cho2023dall, luccioni2023stable,wang2023t2iat,seshadri2023bias,naik2023social}, there is a need for automatic evaluation protocols regarding 1) the \textbf{entire image} and 2) the \textbf{generation process}.
Although previous work has explored bias on the final generated images \cite{luccioni2023stable, bianchi2023easily, cho2023dall, lin2023word}, there is still a lack of analysis about what text-to-image model generates to fill the unguided regions of the image and why it responds distinctly to different gender indicators.

To the best of our knowledge, this is the first work that analyzes the internal components of Stable Diffusion to study where gender bias originated and how it is propagated. We suggest that such disparities arise from the interplay of representational disparities and prompt-image dependencies during image generation: the process involves transitioning from prompt space to image space, potentially treating genders differently and resulting in representational disparities. 
Using template-free natural language prompts, we further study the dependencies between the prompt and the generated image with the inherent cross-attention mechanism, and categorize objects in prompt and image into five dependency groups.
These dependencies/independencies can be modulated by representational disparities. To systematically explore the two intertwined factors, we generate images from a set of (\textit{neutral}, \textit{feminine}, \textit{masculine}) triplet prompts as in Figure \ref{fig:first}, aiming to quantify representational disparities (Sec. \ref{sec:process_evaluation}) and prompt-image dependencies (Sec. \ref{sec:rq3_groups}).

Our evaluation protocol allows us to formulate and answer the following research questions (RQ):

\begin{description}
    \item[RQ1] Do images generated from neutral prompts exhibit greater similarity to those generated from masculine prompts than to images generated from feminine prompts, and if so, why?
    \item[RQ2] Do object occurrences in images significantly vary based on the gender specified in the prompt? If there are differences, do these object occurrences from neutral prompts exhibit greater similarity to those from masculine or feminine prompts?
    \item[RQ3] Does the gender in the input prompt influence the prompt-image dependencies in Stable Diffusion, and if so, which prompt-image dependencies are more predisposed to be affected?
\end{description}

We conduct experiments on three Stable Diffusion models spanning four caption datasets as well as a text set generated by ChatGPT \cite{brown2020language}. Through the generation of triplet prompts with only gender indicators differing, we observe a consistent trend across Stable Diffusion models. Our key findings indicate:

\begin{itemize}
    \item \textbf{For Stable Diffusion, person = man}
    \vspace{-2pt}
        \begin{itemize}
            \item Quantitatively, neutral prompts consistently produce images that look more similar to those from masculine prompts than feminine prompts.
            \item The neutral representations are closer to the masculine representations \textit{for all the internal stages} of the generation process.
        \end{itemize}

    \item \textbf{Explicit objects are consistent across genders}
        \vspace{-2pt}
         \begin{itemize}
        \item Objects generated explicitly from prompts exhibit similar co-occurrence for different genders. 
         \end{itemize}
         
    \item \textbf{Unguided objects are gendered}
        \vspace{-2pt}
        \begin{itemize}
        \item Objects not explicitly mentioned in the prompt are generated at different rates for each gender.
        \item Co-occurrences of objects in images from neutral prompts consistently exhibit greater similarity to those from masculine prompts. 
         \end{itemize}
\end{itemize}
    \vspace{-2pt}
    
Our findings show that gender bias extends beyond people's representations, permeating through the entire image and affecting the generated objects. We conclude the paper with recommendations for both model developers and users, aimed at mitigating this effect.

\begin{table*}[t]
\hspace{-10pt}
\centering
\renewcommand{\arraystretch}{1.1}
\setlength{\tabcolsep}{5pt}
\small
\scalebox{0.93}{
\begin{tabularx}{1.07\textwidth}{@{}l l l c c c c l l l l l@{}}
\toprule
 & \multicolumn{2}{c}{Input} & & \multicolumn{3}{c}{Evaluation space} &  & \multicolumn{1}{c}{Bias}  \\
 \cline{2-3}
 \cline{5-7}
 \cline{9-9}
Method & Prompt type & Prompt variation & & Prompt & Denoising & Image & & Subject of bias\\
\midrule 

\cite{luccioni2023stable} & Template & Identity, Profession & & - & - & \cmark & & Gender \\ 

\cite{bakr2023hrs} & Template & Objects & & - & - & \cmark & & Performance \\ 

\cite{teo2024measuring} & Template & - & & - & - & \cmark & & Gender \\ 

\cite{lee2023holistic} - Fairness & Free-form pairs & Gender & & - & - & \cmark & & Performance \\ 

\cite{lee2023holistic} - Bias & Template & Adjective, Profession & & - & - & \cmark & &  Gender \\ 

\cite{cho2023dall} & Template & Profession & & - & - & \cmark & & Gender, Attire \\ 

\cite{bianchi2023easily} & Template & Profession & & - & - & \cmark & & Gender \\ 

\cite{wang2023t2iat} & Template & Profession & & - & - & \cmark & & Gender  \\ 

\cite{chinchure2023tibet} & Free-form & Gender & & - & - & \cmark & & Gender \\ 

\cite{zhang2023auditing} & Template & Gender, Attire, Activity & & - & - & \cmark & & Attire \\ 

\cite{naik2023social} & Template & Adjective, Profession & & - & - & \cmark & & Gender \\ 
\cite{naik2023social} - Expanded & Template & Gender, Profession & & - & - & \cmark & & Gender, Performance \\ 

\midrule
Ours & Free-form triplets & Gender & & \cmark & \cmark & \cmark & & Layout, Objects  \\ 

\bottomrule
\end{tabularx}}
\caption{Gender bias evaluation methods in text-to-image generation. We compare with previous methods on Input (prompt type, prompt variation), Evaluation space (prompt, denoising, image), and Bias (subject of bias). ``Prompt variation'' refers to how prompts vary in attributes (e.g., profession) while keeping other words unchanged. In terms of the ``Subject of bias'', ``Gender'' means the gender of generated faces, while ``Performance'' contains generation performance metric such as text-to-image alignment and image quality.}
\label{tab:related_work}
\end{table*}

\section{Related work}
\label{sec:related_work}

\paragraph{Text-to-image models}
There are three main types of text-to-image generation models: GANs \cite{goodfellow2020generative, tao2022df, reed2016generative}, autoregressive \cite{ramesh2022hierarchical, ramesh2021zero, ding2021cogview, ding2022cogview2, yu2022scaling}, and diffusion \cite{ho2020denoising, rombach2022high, saharia2022photorealistic}. 
Within diffusion models, Stable Diffusion \cite{rombach2022high} has emerged as the preferred testbed due to its high-quality generations and open-source nature. As diffusion models rely on cross-attention to connect text and image modalities, it enables the examination of the image generation process at the word level \cite{hertz2023prompt}. The cross-attention module assists in tasks such as editing \cite{hertz2023prompt, lu2023tf, epstein2023diffusion, gandikota2023erasing, gandikota2023unified}  and segmentation \cite{tang2022daam, wu2023diffumask, pnvr2023ld}. By leveraging this property, we can investigate the relationship between gender and prompt-guided generations.

\paragraph{Social bias}
Text-to-image generation models often reproduce demographic stereotypes tied to gender and race across various factors, including but not limited to occupations \cite{bianchi2023easily, cho2023dall, luccioni2023stable, wang2023t2iat, mandal2023multimodal, lin2023word, seshadri2023bias}, adjectives \cite{luccioni2023stable, naik2023social, berg2022prompt}, objects \cite{mannering2023analysing}, outfits \cite{zhang2023auditing}, and nationalities \cite{bianchi2023easily, wolfe2022american}. Analysis of prompt templates like ``\texttt{a photo of the face of [OCCUPATION]}'' reveals that certain occupations, such as \textit{software developers}, are predominantly represented as white men, while \textit{housekeepers} tend to be associated with women of color. Additionally, Wolfe et al. \cite{wolfe2023contrastive} showed that models are more inclined to generate sexualized images in response to prompts containing ``\texttt{a [AGE] year old girl}''. Moreover, Zhang et al. \cite{zhang2023iti} argued that unfairness extends to images depicting underrepresented attributes like \textit{wearing glasses}, highlighting the pervasive nature of biases in the generation process. In addition to biases concerning humans, previous studies have explored geographical-level differences in objects \cite{hall2023dig} and the correctness of cultural context \cite{basu2023inspecting, liu2024scoft}.

\paragraph{Bias evaluation}
A fundamental aspect in the study of bias is the evaluation protocol. 
As summarized in Table \ref{tab:related_work}, we compare differences between our method and several previous gender bias evaluation methods in text-to-image generation \cite{luccioni2023stable, bakr2023hrs, teo2024measuring, lee2023holistic, cho2023dall, bianchi2023easily, wang2023t2iat, chinchure2023tibet, zhang2023auditing, naik2023social}.
Most of these approaches rely on prompts that fill attributes (e.g., profession) with a template, leading to constrained scenarios and limited additional details in the prompts. Moreover, these methods evaluate bias on the proxy presentation of the generated images but do not examine presentations in the generation process. Besides, these methods mainly focus on people's attributes, such as the gender of faces, thereby overlooking biases in the generated visual elements as well as the entire image context.
Except for the method that exclusively on gender bias evaluation, there are traditional evaluation criteria for text-to-image models measuring image fidelity and text-image alignment with automated metrics \cite{salimans2016improved, heusel2017gans, vedantam2015cider, papineni2002bleu} or human evaluation \cite{otani2023toward}. 

Overall, there is an absence of automated methods for nuanced bias evaluation that conveys bias at the different stages of the generation process. Using free-form prompts, our work proposes a method to uncover prompt-image dependencies, disclosing how objects are generated differently according to gender indicators in the prompt.
\section{Preliminaries}
\label{sec:preliminaries}

\paragraph{Triplet prompt generation}
\label{para: triplet}
Let $\mathcal{P}_\text{n}$ be a set of \textit{neutral} prompts, which do not specify the gender of the person. As shown in Figure \ref{fig:first}, from these neutral prompts, we generate two counterpart prompt sets, $\mathcal{P}_\text{f}$ and $\mathcal{P}_\text{m}$, as \textit{feminine} and \textit{masculine} prompt sets, respectively. The only difference among these three prompt sets is the gender indicator, while all other words remain unchanged.
Our bias evaluation is based on analyzing distinctions between pairs of generated images from the triplet $\{\mathcal{P}_\text{n}, \mathcal{P}_\text{f}, \mathcal{P}_\text{m}\}$. 

We generate neutral prompts from natural language sentences, consisting of captions from four vision-language datasets (GCC validation set \cite{sharma2018conceptual}, COCO \cite{lin2014microsoft}, TextCaps \cite{sidorov2020textcaps}, and Flickr30k \cite{young2014image}), as well as a profession prompt set generated by ChatGPT 3.5 \cite{brown2020language}.\footnote{Accessed November 2023.} 
From the vision-language datasets, we generate neutral prompts by choosing \textit{neutral captions} that meet two criteria: (1) they contain the word \textit{person} or \textit{people}, and (2) they do not include other words that indicate humans.
To generate feminine and masculine prompts, we swap \textit{person}/\textit{people} in the neutral captions with the gender indicators \textit{woman}/\textit{women} and \textit{man}/\textit{men}, respectively. 
For the profession prompt set, we generate neutral prompts with ChatGPT based on professions, such as \texttt{\small{ecologist}} or \texttt{\small{doctor}}, across $16$ topics.
For example, \texttt{\small{an ecologist studies the ecosystem in a lush green forest}.} To create feminine and masculine prompts, we prepend \textit{female/male} before the profession. Further details can be found in the supplementary materials.

\begin{table}[t]
\centering
\renewcommand{\arraystretch}{1.1}
\setlength{\tabcolsep}{6pt}
\small
\begin{tabularx}{0.88\columnwidth}{l r r r r r}
\toprule
Data & Triplets & Prompts & Seeds & Images \\
\midrule 

GCC (val) & $418$ & $1,254$ & $5$ & $6,270$ \\
COCO & $51,219$ & $153,657$ & $1$ & $153,657$ \\
TextCaps & $4,041$ & $12,123$ & $1$ & $12,123$ \\
Flickr30k & $16,507$ & $49,521$ & $1$ & $49,521$ \\
Profession & $811$ & $2,433$ & $5$ & $12,165$ \\

\bottomrule
\end{tabularx}
\vspace{-5pt}
\caption{Number of generated triplets, prompts, and images for each dataset.}
\label{tab:datasets}
\end{table}

\paragraph{Image generation}
Given prompt $p$ as input, Stable Diffusion transforms it into a text embedding $\mathbf{t}$ in the \textit{prompt} space using the text encoder. This text embedding is fed into the cross-attention module in UNet \cite{ronneberger2015u}, which performs the denoising operations from an initial noise $\mathbf{z}_{T}$ in the latent space.
After $T$ denoising steps, the embedding $\mathbf{z}_{0}$ in the \textit{denoising} space is obtained. Finally, image $x$ in the \textit{image} space is generated from $\mathbf{z}_0$ by the image decoder.  In this work we evaluate Stable Diffusion models: v1.4,\footnote{\url{https://github.com/CompVis/stable-diffusion}} v2.0-base,\footnote{\url{https://huggingface.co/stabilityai/stable-diffusion-2-base}} and v2.1-base\footnote{\url{https://huggingface.co/stabilityai/stable-diffusion-2-1-base}} (denoted as SD v1.4, SD v2.0, and SD v2.1, respectively). The same generation pipeline is used in all the models. 

Table \ref{tab:datasets} reports the details of image generation for each dataset. The seed is the same within each triplet, ensuring the same initial noise $\mathbf{z}_{T}$. To address data scarcity in GCC and Profession sentences, we produce five images per prompt with five different seeds. 
In the following, when mentioning a dataset, we are referring to the generated images whose prompts originate from the corresponding dataset.

\begin{table*}[t]
\renewcommand{\arraystretch}{1.1}
\setlength{\tabcolsep}{6pt}
\small
\centering

\begin{tabularx}{0.95\textwidth}{@{}p{1mm} l  r c r c r r r r r c@{}}
\toprule

& & Prompt & & Denoising & & \multicolumn{6}{c}{Image}\\ 
 \cline{3-3}
 \cline{5-5}
 \cline{7-12}

\multicolumn{2}{l}{Pairs} & $\mathbf{t}$ & & $\mathbf{z}_0$ & & \textit{SSIM} $\uparrow$ & \textit{Diff.~Pix.}$\downarrow$ & \textit{ResNet} $\uparrow$ & \textit{CLIP} $\uparrow$ & \textit{DINO} $\uparrow$ & \textit{split-product} $\uparrow$ \\

\midrule
\rowcolor{oursrow}
\multicolumn{12}{l}{SD v1.4} \\

& (neutral, feminine)  & $0.909$ & & $0.770$ & & $0.516$ & $42.61$ & $0.848$ & $0.794$ & $0.543$ & $0.956$ \\

& (neutral, masculine) & $\textbf{0.931}$ &  & $\textbf{0.798}$ & &  $\textbf{0.543}$ & $\textbf{39.34}$ & $\textbf{0.859}$ & $\textbf{0.808}$ & $\textbf{0.576}$ & $\textbf{0.961}$ \\

\rowcolor{oursrow}
\multicolumn{12}{l}{SD v2.0} \\
& (neutral, feminine)  & $0.980$ & & $0.767$ & & $0.543$ & $39.00$ & $0.847$ & $0.797$ & $0.545$ & $0.957$ \\

& (neutral, masculine)  & $\textbf{0.982}$ & & $\textbf{0.790}$ & & $\textbf{0.571}$ & $\textbf{35.82}$ & $\textbf{0.864}$ & $\textbf{0.817}$ & $\textbf{0.581}$ & $\textbf{0.963}$ \\

\rowcolor{oursrow}
\multicolumn{12}{l}{SD v2.1} \\

& (neutral, feminine)  & $0.980$  & & $0.755$ & & $0.522$ & $41.48$ & $0.842$ & $0.805$ & $0.527$ & $0.952$ \\

& (neutral, masculine)  & $\textbf{0.982}$ & & $\textbf{0.782}$  & & $\textbf{0.552}$ & $\textbf{37.96}$ & $\textbf{0.856}$ & $\textbf{0.820}$ & $\textbf{0.566}$ & $\textbf{0.959}$ \\

\bottomrule
\end{tabularx}
\caption{Representational disparities between neutral, feminine, and masculine prompts in GCC in the three spaces on Stable Diffusion models.}
\label{tab:rq1_representational_disparities_3_versions}
\end{table*}

\paragraph{Gender bias definition}
The interpretation of gender bias varies across literature, resulting in different work attributing different meanings to the term. In this paper, we define gender bias as:
\begin{itemize}
    \item Within the triplet, images generated from \textit{neutral} prompts consistently display greater similarity to those from either \textit{feminine} or \textit{masculine} prompts.
    \item Specific objects tend to appear more frequently in the generated images associated with a specific gender.
\end{itemize}

Whereas objects are not equally distributed in the real world or across cultures, and recognizing that not all disparities regarding genders are inherently problematic (i.e., the association of \textit{dress} with \textit{women} may not be an issue, whereas \textit{kitchen} might), we argue that it is essential to have a methodology for recognizing and quantifying these differences.  Our proposed evaluation protocol is not envisaged to identify objects that perpetuate discrimination and gender stereotypes but to \textit{highlight significant gender disparities}, regardless of whether they are deemed problematic. 

We apply our evaluation protocol in three Stable Diffusion models, and analyze gender bias by addressing our research questions.
\section{Gender disparities in neutral prompts}
\label{sec:rq1}

\paragraph{RQ1} \textit{Do images generated from neutral prompts exhibit greater similarity to those generated from masculine prompts than to images generated from feminine prompts, and if so, why?}

\vspace{10pt}
In this section, we address the above research question through the use of representational disparities.

\subsection{Representational disparities}
\label{sec:process_evaluation}

We use representational disparities to analyze how images generated by different gender indicators compare with respect to neutral prompts. For a given triplet, the analysis consists on comparing the similarity between \textit{neutral} embeddings and \textit{feminine} and \textit{masculine} embeddings. To measure the extent of gender disparities in the generative process, we examine the representational disparities throughout the entire generation, tracking embeddings from the prompt space to the denoising space and the image space, offering insights into when bias is introduced.

\paragraph{Prompt space}
The prompt space is defined as the space in which all text embeddings lie.
Different points in this space provide different semantics to the following image generation process. To measure the disparity between a pair prompt set $\mathcal{P}$ and $\mathcal{P}'$ in the triplet, we compute cosine similarity as

\begin{equation}
    s_\text{P}(\mathcal{P}, \mathcal{P}') = \frac{1}{|\mathcal{P}|}\sum_{p_i, p'_i} \text{cos}(\mathbf{t}, \mathbf{t}'), \label{eq:sim_prompt}
\end{equation}
where $|\cdot|$ is the number of elements in the given set, $\cos(\cdot, \cdot)$ gives cosine similarity, the summation is computed over all prompts $p_i$ from $\mathcal{P}$ and $p'_i$ from $\mathcal{P}'$,\footnote{Subscript $i$ is the index of the prompt to clarify $p_i$ and $p'_i$ are corresponding prompts, derived from the same one.} text embeddings $\mathbf{t}$ and $\mathbf{t}'$ correspond to prompts $p_i$ and $p'_i$, respectively.

\paragraph{Denoising space}
The embedding $\mathbf{z}_0$ after the last denoising process lies in the denoising space. Similarly to the prompt space, we compute cosine similarity as
\begin{equation}
    s_\text{D}(\mathcal{P}, \mathcal{P}') =\frac{1}{|\mathcal{P}|}\sum_{p_i, p'_i} \text{cos}(\mathbf{z}_{0}, \mathbf{z}_{0}') \label{eq:sim_denoising}
\end{equation}
where $\mathbf{z}_{0}$ and $\mathbf{z}_{0}'$ are derived from $p_i$ and $p'_i$, respectively.

\paragraph{Image space}
As bias often involves more in the semantics rather than pixel values, we adopt a spectrum of metrics computed from the generated images. 
To measure image structural differences, we use the average of SSIM scores over all pixels as one of our disparity metrics \textit{SSIM}. Additionally, the ratio of the number of pixels in the contours with higher SSIM scores is used as another disparity metric \textit{Diff.~Pix.}
To quantify differences in higher-level semantics, we apply latent vectors of pre-trained neural networks, adopting the last fully-connected layer of ResNet-50 \cite{he2016deep}, the CLIP image encoder \cite{radford2021learning}, and the last layer of DINO \cite{caron2021emerging}, referred to as \textit{ResNet}, \textit{CLIP}, and \textit{DINO}, respectively. For all metrics, we compute the cosine similarity between the latent vectors from image pairs as in Eq. \ref{eq:sim_prompt} and Eq. \ref{eq:sim_denoising}. Additionally, we adopt \textit{split-product} \cite{somepalli2023diffusion}, computing the maximum cosine similarity among corresponding patches between image pairs.

\subsection{Results analysis} 
\label{sec:rq1}
\begin{table*}[t]
\renewcommand{\arraystretch}{1.1}
\setlength{\tabcolsep}{6pt}
\footnotesize
\centering
\begin{tabularx}{0.55\textwidth}{@{}p{1mm}l r r r r r}
\toprule

\multicolumn{2}{l}{Pairs} & GCC & COCO & TextCaps & Flickr30k & Profession \\ 
\midrule

\rowcolor{oursrow}
\multicolumn{7}{l}{SD v1.4} \\
& $s_{\text{O}}(\mathcal{P}_\text{n}, \mathcal{P}_\text{f})$ & $0.379$ & $0.486$ & $0.413$ & $0.424$ & $0.350$ \\
& $s_{\text{O}}(\mathcal{P}_\text{n}, \mathcal{P}_\text{m})$  & $\textbf{0.414}$ & $\textbf{0.516}$ & $\textbf{0.444}$ & $\textbf{0.457}$ & $\textbf{0.374}$ \\

\rowcolor{oursrow}
\multicolumn{7}{l}{SD v2.0} \\
& $s_{\text{O}}(\mathcal{P}_\text{n}, \mathcal{P}_\text{f})$ & $0.382$ & $0.512$ & $0.420$ & $0.445$ & $0.362$ \\
& $s_{\text{O}}(\mathcal{P}_\text{n}, \mathcal{P}_\text{m})$  & $\textbf{0.425}$ & $\textbf{0.531}$ & $\textbf{0.448}$ & $\textbf{0.476}$ & $\textbf{0.376}$ \\

\rowcolor{oursrow}
\multicolumn{7}{l}{SD v2.1} \\
& $s_{\text{O}}(\mathcal{P}_\text{n}, \mathcal{P}_\text{f})$ & $0.380$ & $0.499$ & $0.388$ & $0.426$ & $0.349$ \\
& $s_{\text{O}}(\mathcal{P}_\text{n}, \mathcal{P}_\text{m})$  & $\textbf{0.419}$ & $\textbf{0.522}$ & $\textbf{0.419}$ & $\textbf{0.451}$ & $\textbf{0.382}$ \\

\bottomrule
\end{tabularx}
\caption{Co-occurrence similarity on Stable Diffusion models.}
\label{tab:rq2_obj_occ_cossim_3_versions}
\vspace{-10pt}
\end{table*}

By analyzing the representational disparities on (\textit{neutral}, \textit{feminine}) and (\textit{neutral}, \textit{masculine}) pairs, we can provide some answers for \textbf{RQ1}.

In the image space, regardless of whether considering the entire image holistically (\textit{SSIM}, \textit{Diff.~Pix}, \textit{ResNet}, \textit{CLIP}, and \textit{DINO}) or the highest similarity on corresponding patches (\textit{split-product}), images generated from \textit{neutral} prompts consistently demonstrate greater similarity to those from \textit{masculine} prompts. Results on GCC-derived prompts are shown in Table \ref{tab:rq1_representational_disparities_3_versions}, whereas results on other datasets and models can be found in the supplementary material. This trend is consistently observed in all datasets and all models. 

Tracing back to the prompt space and denoising space to explore where and when gender bias emerges in the generated images, results in Table \ref{tab:rq1_representational_disparities_3_versions} show that embeddings from \textit{neutral} prompts are closer to the embeddings from \textit{masculine} prompts both in the prompt space and the denoising space. Although Stable Diffusion models apply different text encoders (OpenCLIP-ViT/H for SD v2.0 and SD v2.1, while CLIP ViT-L/14 for SD v1.4), the same trend is observed across all three models and all datasets. This indicates that gender bias originates from the text embedding and perpetuates through the generation process, leading to the disparities observed in the generated images. 

\section{Influence of gender in objects}
\label{sec:rq2}

\begin{figure}[t]
\hspace{-14pt}
\centering
\includegraphics[width=1.05\columnwidth]{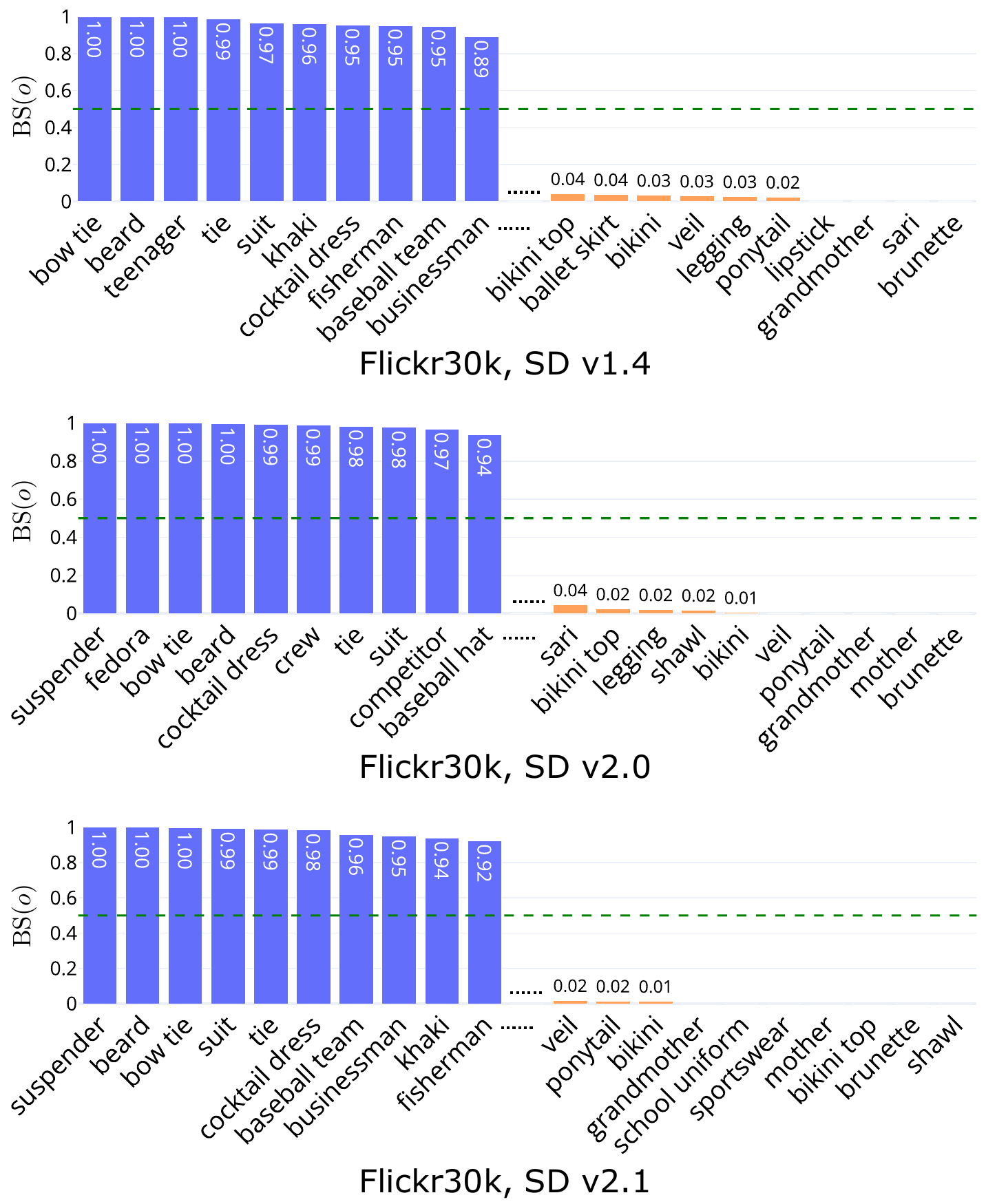}
\vspace{-8pt}
\caption{Bias score in Flickr30k. The higher values (in blue) suggest an object is biased toward masculine prompts, while lower values (in orange) indicate a preference toward feminine prompts. $\text{BS}(o)= 0.5$ (green line) shows the object does not skew toward a certain gender. We filter objects if the maximum co-occurrence is less than $20$.}
\label{fig:rq2_bs}
\end{figure}

\paragraph{RQ2} \textit{Do object occurrences in images significantly vary based on the gender specified in the prompt? If there are differences, do these object occurrences from neutral prompts exhibit greater similarity to those from masculine or feminine prompts?}

\vspace{10pt}
The representational disparities reflect the holistic similarity between gender groups, but they do not convey fine-grained differences, i.e., why a certain object appears in the generated image given a gender-specific prompt. In this section, we address \textbf{RQ2} by investigating the relationship between gender and the objects in the generated images. To do so, we extract objects with a visual grounding model and study their co-occurrence with each gender. 

\subsection{Detecting generated objects}
To detect objects in the generated images we use the assembled model RAM-Grounded-SAM. Given a generated image, RAM \cite{zhang2023recognize} predicts plausibly objects, which are used by Grounded DINO \cite{liu2023grounding} to propose bounding boxes around the candidate objects. Then, Segment Anything Model (SAM) \cite{kirillov2023segany} extracts object regions $m_o$ within the bounding box of the object $o$. For each image, a set of object names and a set of regions are obtained. 

\subsection{Evaluation metrics}
\label{sec:groups_metrics}

Our evaluation protocol involves measuring the differences in object co-occurrences for different genders.
Let $\text{cnt}(o, p)$ denote the number of occurrences of the object $o$ in the image generated from the prompt $p$ in the prompt set $\mathcal{P}$. The total number of co-occurrence $C(o, \mathcal{P})$ is given by:
\begin{equation}
    C(o, \mathcal{P}) = \sum_{p \in \mathcal{P}} \text{cnt}(o, p) \label{eq:count}
\end{equation}
With the above definition and a set of triplet prompts, we use the following three methods to evaluate the influence of gender in the generated objects:

\paragraph{1) Statistical tests}
We use the chi-square test to check whether there are statistical differences in the object co-occurrence among two or three image sets. 
This test is applicable to the triplet and any pairs in the triplet. If the resulting $p$-value is below $0.05$, we interpret significant differences in the object distribution in the pair or triplet.

\paragraph{2) Co-occurrence similarity}
We compute the similarity of the co-occurrences of detected objects between two image sets. Formally, let the vector $\mathbf{v}_p$ denote the object occurrences in the image generated from prompt $p$, and each element in $\mathbf{v}_p$ is the occurrence $\text{cnt}(o, p)$ for the object $o$ in the image. Similarly to Eq. \ref{eq:sim_prompt} and Eq. \ref{eq:sim_denoising}, we compute cosine similarity on object co-occurrences as
\begin{equation}
    s_{\text{O}}(\mathcal{P}, \mathcal{P}') =\frac{1}{|\mathcal{P}|}\sum_{p_i, p'_i} \text{cos}(\mathbf{v}_i, \mathbf{v}'_i) \label{eq:sim_obj},
\end{equation}
where prompt sets $\mathcal{P}$ and $\mathcal{P}'$ are in the triplet. $\mathbf{v}_i$ and $\mathbf{v}'_i$ are derived from prompt $p_i$ in $\mathcal{P}$ and $p'_i$ in $\mathcal{P}'$, respectively. A higher co-occurrence similarity means that objects are detected with the same-level frequency in two image sets, whereas a low similarity means that objects are detected at different rates.

\paragraph{3) Bias score}
Following \cite{zhao2017men}, we compute the bias score $\text{BS}(o)$ for a certain object $o$ as:
\begin{equation}
    \text{BS}(o) = \frac{C(o, \mathcal{P}_\text{m})}{C(o, \mathcal{P}_\text{m}) + \frac{|\mathcal{P}_\text{m}|}{|\mathcal{P}_\text{f}|}C(o, \mathcal{P}_\text{f})}.
\end{equation}
$\text{BS}(o)$ ranges from $0$ to $1$, with $1$ meaning the object is skewed towards \textit{masculine} prompts and $0$ towards \textit{feminine} prompts. If $\text{BS}(o)= 0.5$, object $o$ does not favor any gender.

\subsection{Results analysis}

All the $p$-values from chi-square tests among the triplets and pairs are below $10^{-5}$, implying significant differences in the object distributions of each gender across all datasets and models. This shows that according to gender, not only the person in the image may change, but also the objects generated in the image are statistically different.

To investigate whether the object co-occurrences 
of neutral images exhibit larger similarity to a certain gender image set, we compute co-occurrence similarity on pairs (\textit{neutral, feminine}) and (\textit{neutral, masculine}).
Results in Table \ref{tab:rq2_obj_occ_cossim_3_versions} indicate that object co-occurrences in \textit{neutral} consistently exhibit greater similarity to those in \textit{masculine} prompts than in \textit{feminine} prompts across all datasets and models, corroborating the observations in Section \ref{sec:rq1}. This, again, indicates that prompts that use gender neutral words tend to generated objects that are more commonly generated for masculine prompts than for feminine prompts.

Subsequently, we examine specific examples by computing the bias score for each object in the generated images. Figure \ref{fig:rq2_bs} shows results on Flickr30k. Results for other datasets and models can be found in the supplementary material. 
We can observe that objects with higher or lower bias scores in different versions of Stable Diffusion show a similar pattern.
Thus, we analyze results on SD v2.0 as an example. Notably, clothing exhibits a high bias: for example \texttt{suspender}($1$), \texttt{fedora}($1$), and \texttt{bow tie}($1$) lean towards \textit{masculine}, while \texttt{veil}($0$), \texttt{bikini}($0.01$), and \texttt{shawl}($0.02$) lean towards \textit{feminine}. This is not surprising, considering that clothing elements are traditionally gendered.
Other than clothing, we find a strong association between \texttt{family}($0.11$) and \texttt{child}($0.31$) with \textit{feminine} prompts, potentially associating \textit{feminine}  with caregiver, while \textit{masculine} prompts exhibit greater alignment with words related to sports such as \texttt{baseball team}($0.91$), \texttt{skateboarder}($0.89$), and \texttt{golfer}($0.86$), a phenomenon that has been previously observed in VQA datasets \cite{hirota2022gender}.
Another observation is that \textit{feminine} prompts also have a high association with food, such as \texttt{salad}($0.22$), \texttt{meal}($0.25$), and \texttt{cotton candy}($0.31$). 
Results on other versions of Stable Diffusion and other datasets show similar trends, and additionally reveal that \texttt{businessman} tends to be skewed towards \textit{masculine} whereas \texttt{kitchenware} tends to be associated with \textit{feminine} prompts.

\section{Gender in prompt-image dependencies}
\label{sec:rq3_groups}
\paragraph{RQ3} \textit{Does the gender in the input prompt influence the prompt-image dependencies in Stable Diffusion, and if so, which prompt-image dependencies are more predisposed to be affected?}

\begin{figure}[t]
\centering
\includegraphics[width=0.98\columnwidth]{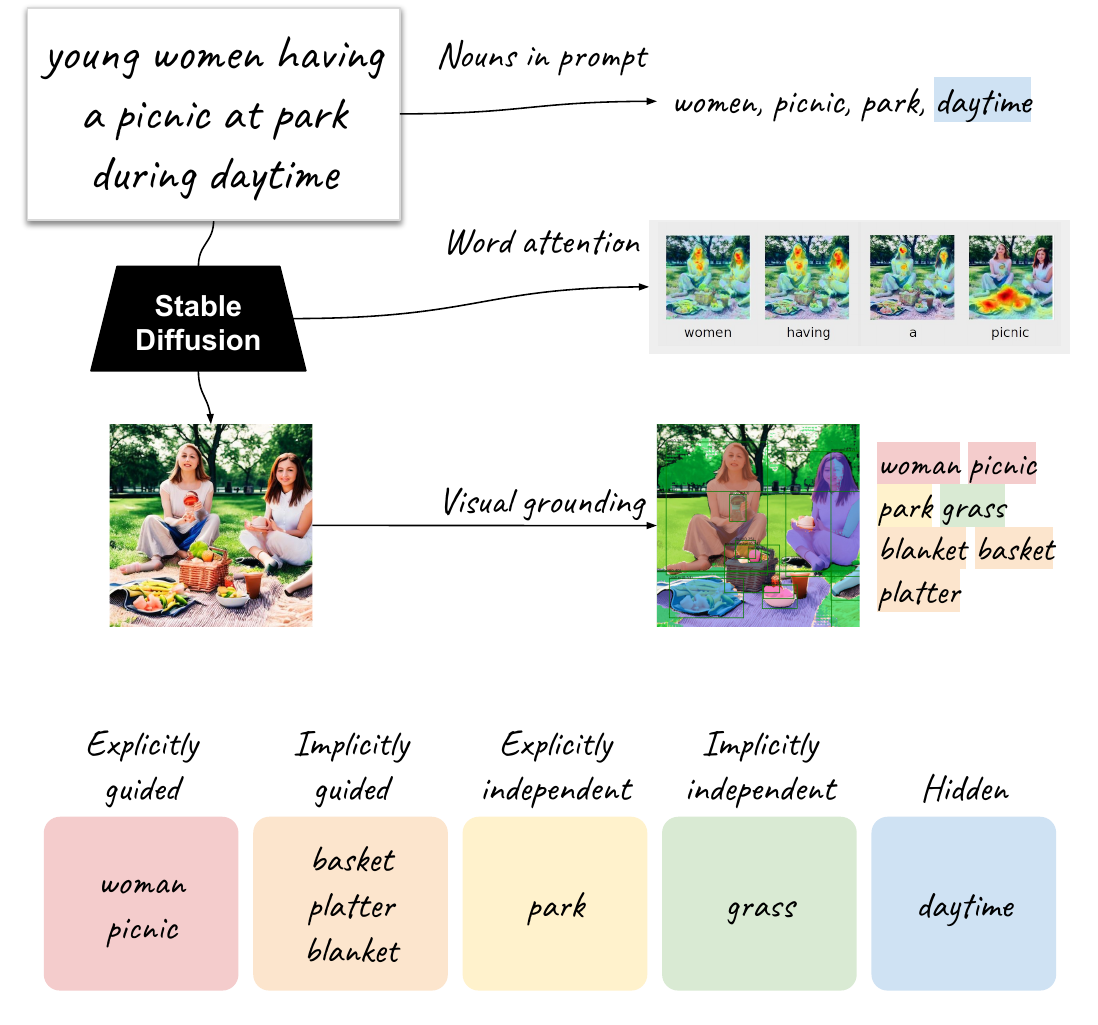}
\vspace{-5pt}
\caption{Prompt-image dependency groups.}
\label{fig:groups_example}
\end{figure}

\vspace{10pt}
To answer to this question, we need to know not only which objects are generated for each gender but also how each object is generated in the diffusion process. To do so, we propose to classify objects into prompt-image dependency groups according to their relationship with the input prompt and the generated image. First, we conduct an \textit{extended object extraction} by detecting not only the objects in the generated image as in Section \ref{sec:rq2}, but other objects also involved in the generative process. Then, we classify each object according to five \textit{prompt-image dependency groups}, which allows us to study how gender influences objects according to their generative process.

\subsection{Extended object extraction}
To detect extended objects involved in the generative process, we conduct three extraction processes.

\paragraph{1) Nouns in prompt}
Prompts, designed by users, are a direct cue of what they wish to see in the generated image. The generated image, on the other hand, is required to be faithful to the prompt. The first extraction process targets nouns within the prompt, recognizing their importance in directly shaping the occurrence of objects in the generated image. For each prompt, we obtain a noun set including all lemmatized nouns $n$ in the prompt by using NLTK \cite{bird2009natural}.

\paragraph{2) Word attention}
\label{para:word_attn}
Verifying whether objects in the noun set are faithfully generated in the image is demanding, as it requires locating the region where the noun guides.
Fortunately, cross-attention has proven to be effective in exploring the word guidance during the generation process \cite{hertz2023prompt, tang2022daam}. Our second extraction process is the word attention masks generated by the cross-attention module via DAAM \cite{tang2022daam}. 
For each word, we first compute the normalized attention map,\footnote{Details are in the supplementary material.} where a higher value indicates that the pixel is more associated with the word. Then, we binarize the attention map with a threshold $\theta$ to obtain a set of masks $a_n$, responding to the region of an object specified by the word $n$. In each prompt, we obtain a mask set containing the mask $a_n$ for each word $n$.

\paragraph{3) Visual grounding}
Nouns and the corresponding object regions cover only a small subset of objects in the generated image; there should be many other objects that are not explicitly described in the prompt but are still included in the image to complete the scene. We aim to enumerate as many objects as possible for comprehensive object-level analysis. To spot regions of arbitrary objects, the last extraction process is the same visual grounding process as in Section \ref{sec:rq2}.


\subsection{Prompt-image dependency groups}
\label{sec:object_groups}

Next, we classify each detected object according to its generative process. On one hand, the generated image should align with its prompt, which can be verified using the noun set and the mask set. On the other hand, the image may have other visual elements beyond the prompt, listed in the object set and the object region set. To define prompt-image dependency groups, we consider the dependency among objects, the noun set, and the mask set based on its membership. 

\begin{definition}[Explicitly]
If the object $o$ is in the noun set, it is \textit{explicitly} described in the prompt.
\end{definition}

\begin{definition}[Guided]
If object region $m_o$ \textit{sufficiently} overlaps with at least one mask in the mask set, the object $o$ is \textit{guided} by cross-attention between the prompt and the image. Sufficiency is determined by the coverage of object region $m_o$ by the mask $a$: 
\begin{equation}\label{eq:coverage}
    \text{coverage}(m_o, a) = \frac{|m_o \cap a|}{|m_o|},
\end{equation}
where $|\cdot|$ is the number of pixels. Thus, if $\text{coverage}(m_o, a)$ is larger than a certain threshold $\sigma$, the object region $m_o$ sufficiently overlaps with the mask $a$.
\end{definition}
\label{para:sufficiently_overlap}

\begin{figure*}[t]
\hspace{-8pt}
\centering
\includegraphics[width=0.94\textwidth]{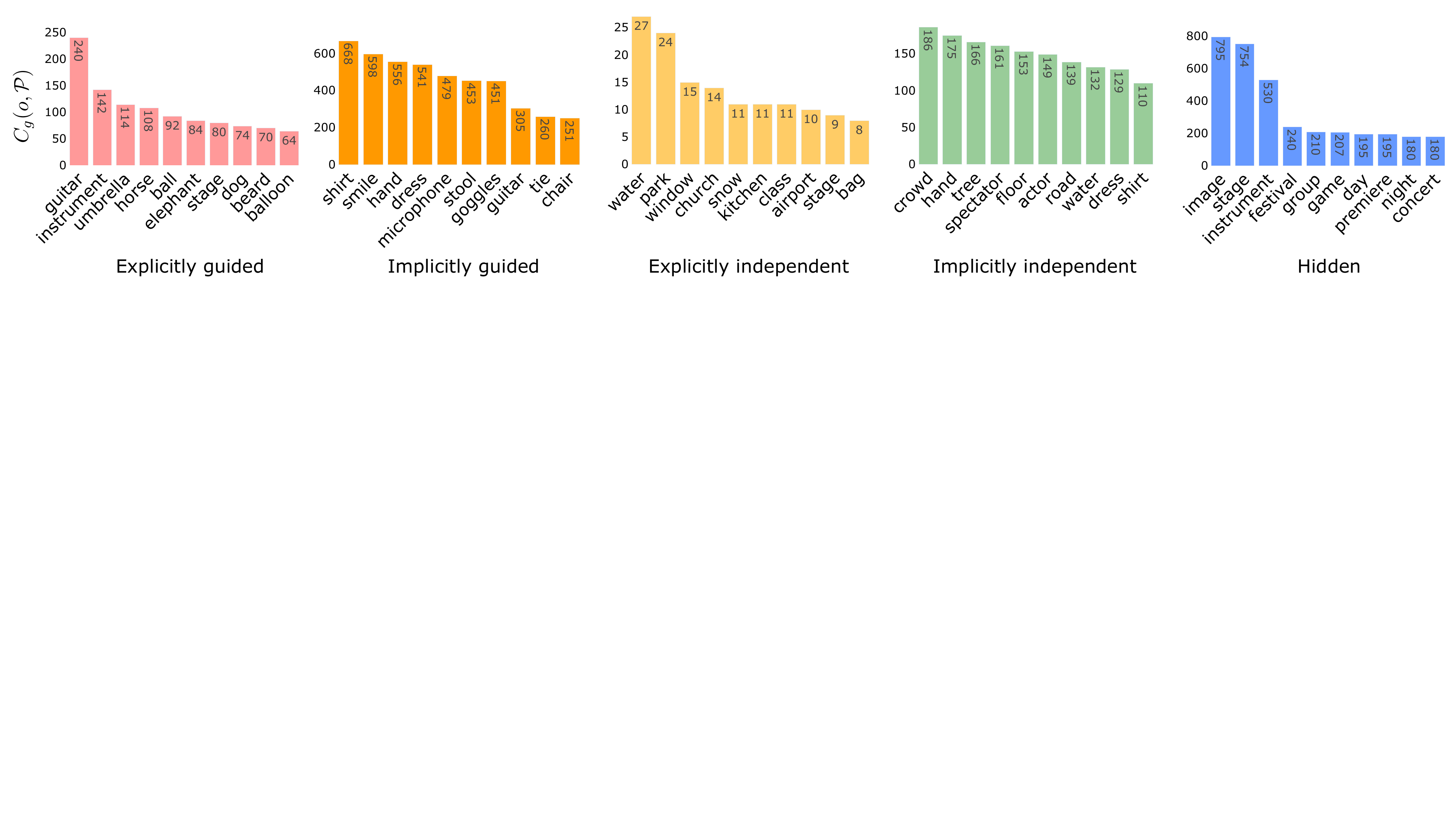}
\caption{Top-$10$ most frequent objects in each prompt-image dependency group in GCC on SD v2.0.}
\label{fig:obj_in_group}
\end{figure*}

With these definitions, we cluster objects in the object set into five groups, as illustrated in Figure \ref{fig:groups_example} with the example prompt \texttt{\small{young women having a picnic at the park during daytime}}:

\begin{description}
    \item[\colorbox{g1}{Explicitly guided}] The object is \textit{explicitly} mentioned in the prompt and \textit{guided} by cross-attention. Faithful image generation may require each noun to be associated with the corresponding object.
    
    \item[\colorbox{g2}{Implicitly guided}] The object is \textit{not explicitly} mentioned in the prompt but \textit{guided} by cross-attention. The object may be strongly associated with or pertain to a certain noun in the noun set, e.g., the object \texttt{\small{basket}} for the noun \texttt{\small{picnic}}.

    \item[\colorbox{g3}{Explicitly independent}] The object is \textit{explicitly} mentioned in the prompt but \textit{not guided} by cross-attention. e.g., \texttt{\small{park}}.
    
    \item[\colorbox{g4}{Implicitly independent}] The object is \textit{not explicitly} mentioned in the prompt and \textit{not guided} by cross-attention. The object is generated solely based on contextual cues, e.g., $\texttt{\small{grass}}$.

    \item[\colorbox{g5}{Hidden}] The noun has no association with objects in the object set, i.e., the noun is \textit{not included} in the images, e.g., $\texttt{\small{daytime}}$.
\end{description}

Figure \ref{fig:groups_example} illustrates the object extraction processes and the resulting dependency groups. Dependency groups are important as they depict if an object tends to appear, for example, in relation to the prompt (\colorbox{g1}{explicitly guided}) or just for filling the scene (\colorbox{g4}{implicitly independent}). Together with the gender-specific sets of prompts, they vividly provide essential insights into how an image generation model behaves for different genders.

\subsection{Result analysis} 
We denote co-occurrence $C_g(o, \mathcal{P})$ as the number of occurrences of object $o$ in each dependency group $g$. To clarify, given that there are nouns included in the hidden group, the computation of occurrence should be adjusted from $C_g(o, \mathcal{P})$ to $C_g(n, \mathcal{P})$ for $n$ in the hidden group. 

\paragraph{Objects in dependency groups}
To answer RQ3, we first investigate objects in the prompt-image dependency groups, aiming to identify which types of objects are generated under the influence of the prompt, the cross-attention, or the context of the generated image. As shown in Figure \ref{fig:obj_in_group}, we look into the prevalent objects within each dependency group on SD v2.0.\footnote{To focus on the differences between generated objects, we remove individuals (person, people, women, woman, men, man, female, male, girl, boy).} Although the specific generated objects align with the prompt's domain, and their frequencies may vary across datasets, we observe consistent trends. 

Objects in the \colorbox{g1}{explicitly guided} group include animals and tangible items commonly encountered in daily life, such as \texttt{guitar} and \texttt{umbrella}. The \colorbox{g2}{implicitly guided} group contains objects surrounding human beings, such as clothing and personal belongings like \texttt{shirt} and \texttt{microphone}. The \colorbox{g3}{explicitly independent} group comprises words related to the surrounding environment, such as \texttt{park} or \texttt{church}. Objects in the \colorbox{g4}{implicitly independent} group are typically part of the background that can be detected, like \texttt{crowd} and \texttt{tree}, along with attire accompanying individuals. Lastly, the \colorbox{g5}{hidden} group comprises words challenging to detect in images, such as \texttt{game} and \texttt{day}.

\begin{figure*}[t]
\hspace{-16pt}
\centering
\includegraphics[width=0.88\textwidth]{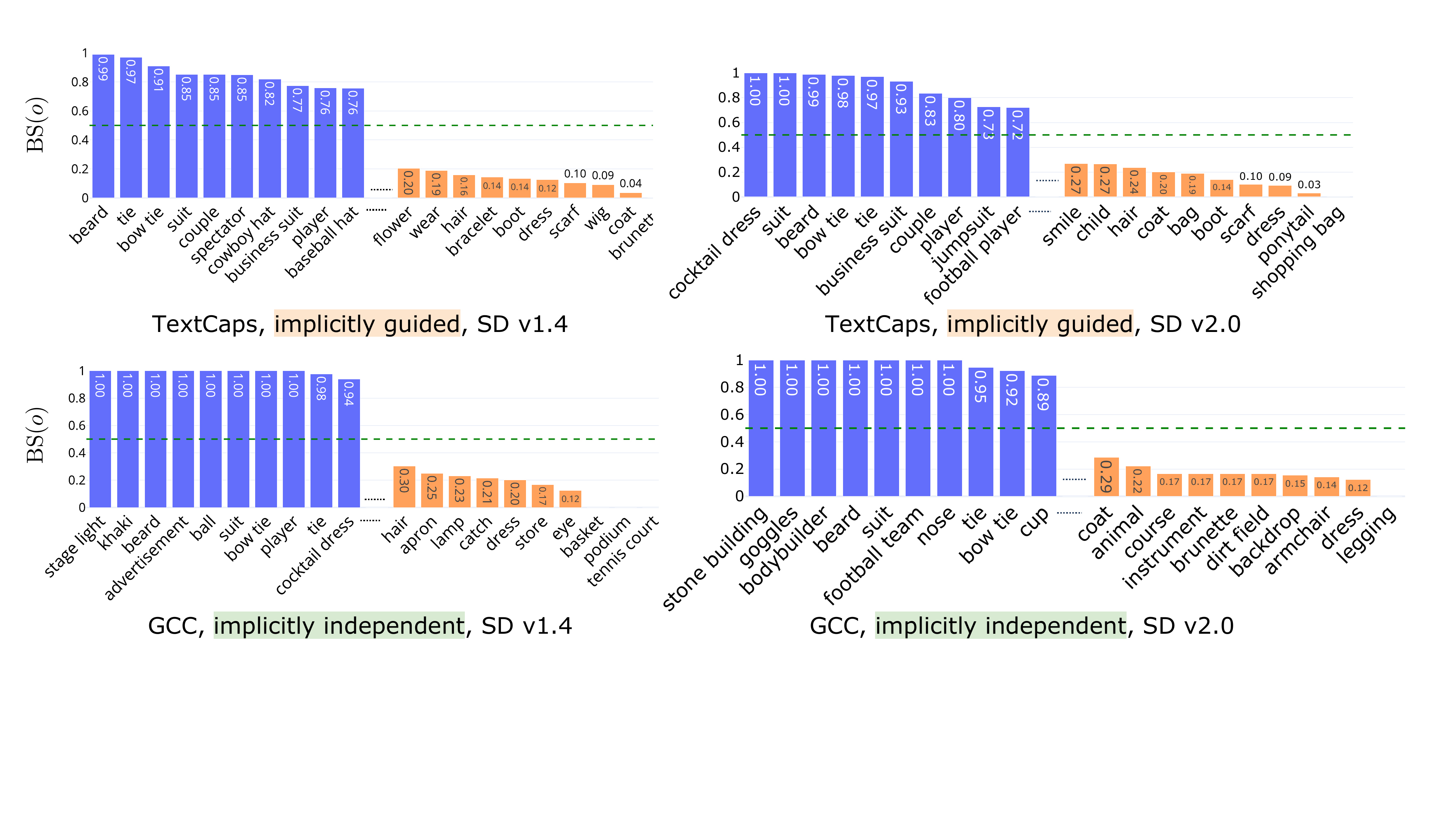}
\caption{Bias score by groups on SD v1.4 and SD v2.0. Top: \colorbox{g2}{implicitly guided} group in TextCaps on SD v1.4 and SD v2.0. Bottom:  \colorbox{g4}{implicitly independent} group in GCC on SD v1.4 and SD v2.0.}
\label{fig:rq3_bias_score_g2_g4}
\end{figure*}

\paragraph{Gender and dependency groups}
Next, we investigate the relationship between gender and the objects in each prompt-image dependency group. 
To discern whether object differences are statistically significant, we conduct chi-square tests\footnote{Details and results can be found in the supplementary material.} on the object co-occurrence for each dependency group. 
While we find significant differences ($p\text{-value}<0.05$) across all datasets in the \colorbox{g2}{implicitly guided} and \colorbox{g4}{implicitly independent} groups, we do not find significant differences in most datasets in the \colorbox{g1}{explicitly guided}, \colorbox{g3}{explicitly independent} and \colorbox{g5}{hidden} groups. This suggests that while Stable Diffusion may consistently generate the nouns explicitly mentioned in the prompt, it may rely on gender cues for generating elements that are not specified in the prompt, such as the background and surroundings of the individuals. 

To explore further into the text-image dependencies and their correlation with gender, we calculate the bias score on object co-occurrence in the \colorbox{g2}{implicitly guided} and the \colorbox{g4}{implicitly independent} groups, both of which exhibit statistically significant differences. Figure \ref{fig:rq3_bias_score_g2_g4} shows the top-$10$ objects skewed toward \textit{masculine} and \textit{feminine} in TextCaps and GCC datasets on SD v1.4 and SD v2.0.\footnote{Results on other datasets and models are in the supplementary material.} We filter objects if maximum co-occurrence is less than $20$ in TextCaps, and $5$ in GCC. 
We analyze results on SD v2.0 as examples. For the \colorbox{g2}{implicitly guided} group in TextCaps, we observe high bias scores for clothing items, such as \texttt{cocktail dress}($1$), \texttt{suit}($1$), \texttt{bow tie}($0.98$), and \texttt{tie}($0.97$) for \textit{masculine} and \texttt{ponytail}($0.03$), \texttt{dress} ($0.09$) and \texttt{boot}($0.14$) for \textit{feminine}, aligning with observations in previous work \cite{zhang2023auditing}. 
Another prominent observation, consistent with the findings on Flickr30k in RQ2, is the strong association of \texttt{child}($0.27$) with \textit{feminine}, and \textit{masculine} with sports-related terms such as \texttt{player}($0.8$) and \texttt{football player}($0.72$). Similar gendered associations are observed across different datasets and models.
For the \colorbox{g4}{implicitly independent} group in GCC, words related to sports such as \texttt{bodybuilder}($1$) and \texttt{football team}($1$) are again skewed toward \textit{masculine}, while \texttt{instrument}($0.17$) and \texttt{apron}($0.33$) are skewed to \textit{feminine}. There are also disparities in the words indicating backgrounds, such as \texttt{backdrop}($0.15$) and \texttt{dirt field}($0.17$) for \textit{feminine} and \texttt{stone building}($1$) and \texttt{tennis court}($0.63$) for \textit{masculine}.
Other datasets and models report similar results. Furthermore, it is observed that \texttt{smile} and \texttt{flower} are skewed towards \textit{feminine}.

\section{Additional experiments}
\label{sec:analysis}


To further evaluate our protocol, we conduct intro-prompt evaluation and human evaluation.

\subsection{Intra-prompt evaluation}
To eliminate the influence of randomness, we investigate the research questions using images generated from the same triplet prompts. We generate a total of $3,000$ images on $1,000$ seeds with SD v2.0, from triplet prompts derived from a caption in GCC: ``\textit{person looks at the falling balloons at the conclusion}''. We use the same settings as conducted in the experiments above. 

For RQ1, results provided in the supplementary material show that neutral is consistently closer to masculine across all spaces.
For RQ2, the chi-square tests on the object occurrences among the triplets and every pair within the triplets, $p$-value is consistently less than $\num{e-5}$, indicating statistically significant differences. 
For RQ3, the chi-square tests also reveal significant differences in the groups \colorbox{g2}{implicitly guided} and \colorbox{g4}{implicitly independent} ($p < \num{e-5}$). However, we do not apply chi-square test on \colorbox{g1}{explicitly guided}, \colorbox{g3}{explicitly independent}, and \colorbox{g5}{hidden}, as the numbers of objects in these groups are less than $5$.
The co-occurrence similarity $s_{\text{O}}(\mathcal{P}_\text{n}, \mathcal{P}_\text{f})$ between the neutral and feminine is $0.733$, while the similarity $s_{\text{O}}(\mathcal{P}_\text{n}, \mathcal{P}_\text{m})$ between the neutral and masculine is $\textbf{0.773}$.
This indicates that the object co-occurrences in images generated from \textit{neutral} prompts are closer to those from \textit{masculine} prompts than \textit{feminine} prompts. These findings correspond to the above results.

\subsection{Human evaluation}
To evaluate the reliability of the visual grounding model, we randomly select $100$ generated images from SD v2.0 along with the nouns from the corresponding prompts and conduct a human evaluation to determine whether the nouns are present in the images. The $100$ prompts contain $346$ nouns, from which $227$ ($65.61\%$) are correctly identified both by humans and the automated vision grounding. Out of the remaining $119$ nouns, only $8$ nouns are detected by the model but not observed by humans. These nouns are \texttt{frisbee}(2), \texttt{women}(1), \texttt{people}(1), \texttt{kite}(1), \texttt{scooters}(1), \texttt{tennis}(1) and \texttt{speaker}(1). For the nouns not detected by the model but identified by humans, the most frequent ones are \texttt{woman}(10), \texttt{street}(7), \texttt{people}(6), and \texttt{snowy}(4). The absence of the noun \texttt{street} in the model's detection might be attributed to the strict alignment between nouns and objects. Even if the model successfully identifies \texttt{street scene}, the specific noun \texttt{street} might be placed in one of the \colorbox{g2}{implicitly guided}, \colorbox{g4}{implicitly independent}, or \colorbox{g5}{hidden} groups. 
These results indicate that the visual grounding model has reasonable accuracy in detecting nouns appearing in the generated images, though there is still room for improvement on abstract nouns and scene-level nouns.

\section{Recommendations}
Our methodology revealed significant disparities in the objects generated by three Stable Diffusion models according to the gender in the input prompt. While these discrepancies may seem harmless, they can potentially reinforce gender stereotypes. With this in mind, we propose a series of suggested practices aimed at mitigating these concerns, both for model developers and for users:

\subsection{Model developers}

\paragraph{Debias text embeddings}
We have identified that gender bias originates in the text embedding, with \textit{neutral} prompts consistently being more similar to \textit{masculine} prompts than to \textit{feminine} prompts, and propagates through the entire generation process. Given the documented presence of gender bias in CLIP \cite{wolfe2023contrastive, wolfe2022markedness, agarwal2021evaluating, wolfe2022evidence}, it comes as no surprise that text-to-image generation models relying on CLIP also exhibit such biases.  The first mitigation technique should focus on debiasing the text embedding space, aiming for more equitable representations. 

\paragraph{Identify problematic representations}
While some associations of certain objects with specific genders may not immediately raise concerns, others could potentially do so. Therefore, researchers must meticulously assess these associations, taking into account the cultural context in each instance. It is crucial to examine the co-occurrence of objects across genders and check whether neutral prompts tend to exhibit a preference toward a particular gender.

\paragraph{Investigate modules that complete the scene}
Significant differences were observed in the \textit{implicitly} generated objects, underscoring the need to investigate how the model completes the scene. Future research could explore other modules, probing fine-grained control over the regions not guided by the input. 

\subsection{Users}

\paragraph{Explicitly specify objects}
Our results showed that there are no significant differences in the objects explicitly mentioned in the input prompts concerning gender. This suggests that Stable Diffusion models can adhere to the simple instructions in the prompt regardless of gender.  Therefore, expanding the number of objects in the input could offer greater control over broader guided regions and potentially lead to the generation of images with less gender disparity.

\paragraph{Explicitly specify gender}
Considering that \textit{neutral} prompts consistently produced images more similar to those from \textit{masculine} prompts, we advise refraining from using neutral prompts if targeting a balanced distribution across genders. Instead, using prompts with specified gender indicators may be more reliable.

\section{Limitations}
We acknowledge that our proposed evaluation protocol has limitations, and we emphasize them here for transparency and to inspire the community to propose enhancements in future studies. Firstly, our evaluation protocol focuses on binary genders, neglecting to evaluate gender from a broader spectrum perspective. To enhance inclusivity, future research could extend the analysis to encompass a more diverse range of genders.
Secondly, our protocol relies on a stringent alignment between nouns and objects, assuming their identity after lemmatization, which may overlook variations and synonyms.
Thirdly, the objects segmented in visual grounding may encounter errors, possibly perpetuating issues in the classified groups.
Additionally, if gender bias exists in the visual grounding model, where certain objects may be more challenging to detect in specific genders, this bias could transfer to the final results.
Besides, when the object comprises more than one word (e.g., ``picnic basket''), each noun in the phrase has its own word attention rather than being considered as a single entity.
Last but not least, our study only examines the presence of objects not differentiating with distinct attributes, such as color or shape.

\section{Conclusion}
We introduced an automated evaluation protocol to study gender bias in image generation by probing the internal components of Stable Diffusion models. We investigated both representational disparities and prompt-image dependencies to uncover the origin of bias and how it manipulated image generation. Through the generation of free-form triplet prompts with only gender indicators differing, our findings indicate that:

\begin{enumerate}
    \item Prompts that use \textit{neutral} words to refer to people (\texttt{a person in a park)} consistently yield images more similar to the ones generated from prompts with \textit{masculine} words  (\texttt{a man in a park)} than from prompts with \textit{feminine} words (\texttt{a woman in a park)}.
    \item There are statistically significant differences in the type of objects generated in the image based on the gender indicators in the prompt.
    \item The frequency of objects generated explicitly from prompts exhibit similar behavior for different genders. 
    \item Objects not explicitly mentioned in the prompt exhibit significant differences for each gender.
    \item We particularly observed significant statistical disparities in generated objects based on gender in items related to clothing and traditional gender roles such as sports, which are highly skewed towards images generated from \textit{masculine} prompts, and food, which are skewed towards images generated from \textit{feminine} prompts.
\end{enumerate}

Based on these observations, we provided recommendations for developers and users to reduce such representational disparities and gender bias in the generated images.
We hope these insights contribute to underscoring the nuanced dynamics of gender bias in image generation, offering a new and valuable perspective to the growing body of research on this topic.

\paragraph{{Acknowledgments}}
This work is partly supported by JST CREST Grant No. JPMJCR20D3, JST FOREST Grant No. JPMJFR216O, JSPS KAKENHI Nos. JP22K12091 and JP23H00497, JST SPRING Grant No. JPMJSP2138.

\bibliography{aaai24}

\appendix
\clearpage
\renewcommand{\thetable}{S\arabic{table}}  
\renewcommand{\thefigure}{S\arabic{figure}}
\setcounter{table}{0}
\setcounter{figure}{0}
\setcounter{page}{1}
\setcounter{footnote}{0}

\appendix

This supplementary material provides further experiment details and results related to the three research questions discussed in the main paper. The document is organized into the following sections:
\begin{itemize}[noitemsep,topsep=0pt]
    \item Section \ref{sec:sup:details}: Experiment details.
    \item Section \ref{sec:sup:rq1}: Research question 1.
    \item Section \ref{sec:sup:rq2}: Research question 2.
    \item Section \ref{sec:sup:rq3}: Research question 3.
    \item Section \ref{sec:sup:intra}: Intra-prompt evaluation.
    \item Section \ref{sec:sup:groups_analysis}: Dependency groups analysis.
\end{itemize}

\section{Experiment details}
\label{sec:sup:details}
\subsection{Details}
\paragraph{Model details}
In Sec. \ref{sec:process_evaluation}, we use CLIP ViT-B/32\footnote{\url{https://github.com/openai/CLIP}} \cite{radford2021learning} for \textit{CLIP}. 
For \textit{DINO}, we use DINO-s16 \cite{caron2021emerging} following \cite{ruiz2023dreambooth}. For split-product \cite{somepalli2023diffusion}, we use DINO-b8 \cite{caron2021emerging} following the default configuration.
We utilize RAM-Grounded-SAM\footnote{\url{https://github.com/IDEA-Research/Grounded-Segment-Anything}} \cite{liu2023grounding, kirillov2023segany, zhang2023recognize} for obtaining visual grounding, incorporating RAM (14M) \cite{zhang2023recognize}, GroundingDINO-T \cite{liu2023grounding}, and ViT-H SAM model \cite{kirillov2023segany}.

\paragraph{Thresholds}
In Sec. \ref{para:word_attn}, the threshold $\theta$ for obtaining mask from attention maps is set at $0.35$. 

To determine \textit{sufficient} overlap between word attention and visual grounding (as described in Sec. \ref{para:sufficiently_overlap}), we set different values for threshold $\sigma$ for words referring to humans\footnote{people, person, woman, women, man, men.} versus objects. The reason is that for human words, the generated people in the images can still be considered as \textit{guided} by those words even when the coverage(Eq. \ref{eq:coverage}) of word attention and visual grounding is relatively low. For example, word attention on human words may focus only on the face, while the visual grounding covers the whole body. Since these partial overlap cases are common for human words, we set a lower threshold of $\sigma = 0.25$ when both the detected object and word refer to humans. For all other cases, a higher threshold of $\sigma = 0.7$ is used.

\begin{table}[t]
\renewcommand{\arraystretch}{1.2}
\setlength{\tabcolsep}{7pt}
\small
\centering

\begin{tabularx}{1\columnwidth}{@{}p{0.15\columnwidth}  p{0.8\columnwidth}}
\toprule
Type & Word \\
\midrule

Gender & woman, female, lady, mother, girl, aunt, wife, actress, princess, waitress, sister, queen, pregnant, daughter, she, her, hers, herself, bride, mom, queen, man, male, father, gentleman, boy, uncle, husband, actor, prince, waiter, son, brother, guy, emperor, dude, cowboy, he, his, him, himself, groom, dad, king \\
\hline
Geography & American, Asian, African, Indian, Latino \\
\hline
Others & commander, officer, cheerleader, couple, player, magician, model, entertainer, astronaut, artist, student, politician, family, guest, driver, friend, journalist, relative, hunter, tourist, chief, staff, soldier, civilian, author, prayer, pitcher, singer, kid, groomsman, bridemaid, ceo, customer, dancer, photographer, teenage, child, u, me, I, leader, crew, athlete, celebrity, priest, designer, hiker, footballer, hero, victim, manager, Mr, member, partner, myself, writer \\

\bottomrule
\end{tabularx}
\caption{Words that indicate humans.}
\label{tab:supp_words}
\end{table}

\subsection{Prompts}
\paragraph{Words that indicate humans}
Table \ref{tab:supp_words} displays words that indicate humans. In the four caption datasets (GCC \cite{sharma2018conceptual}, COCO \cite{lin2014microsoft}, TextCaps \cite{sidorov2020textcaps}, and Flick30k \cite{young2014image}), to ensure the sentences do not contain other words that might potentially define the gender of generated people, we filter out sentences containing words listed in Table \ref{tab:supp_words} and their plurals.

\begin{figure*}[t]
\hspace{-8pt}
\centering
\includegraphics[width=1\textwidth]{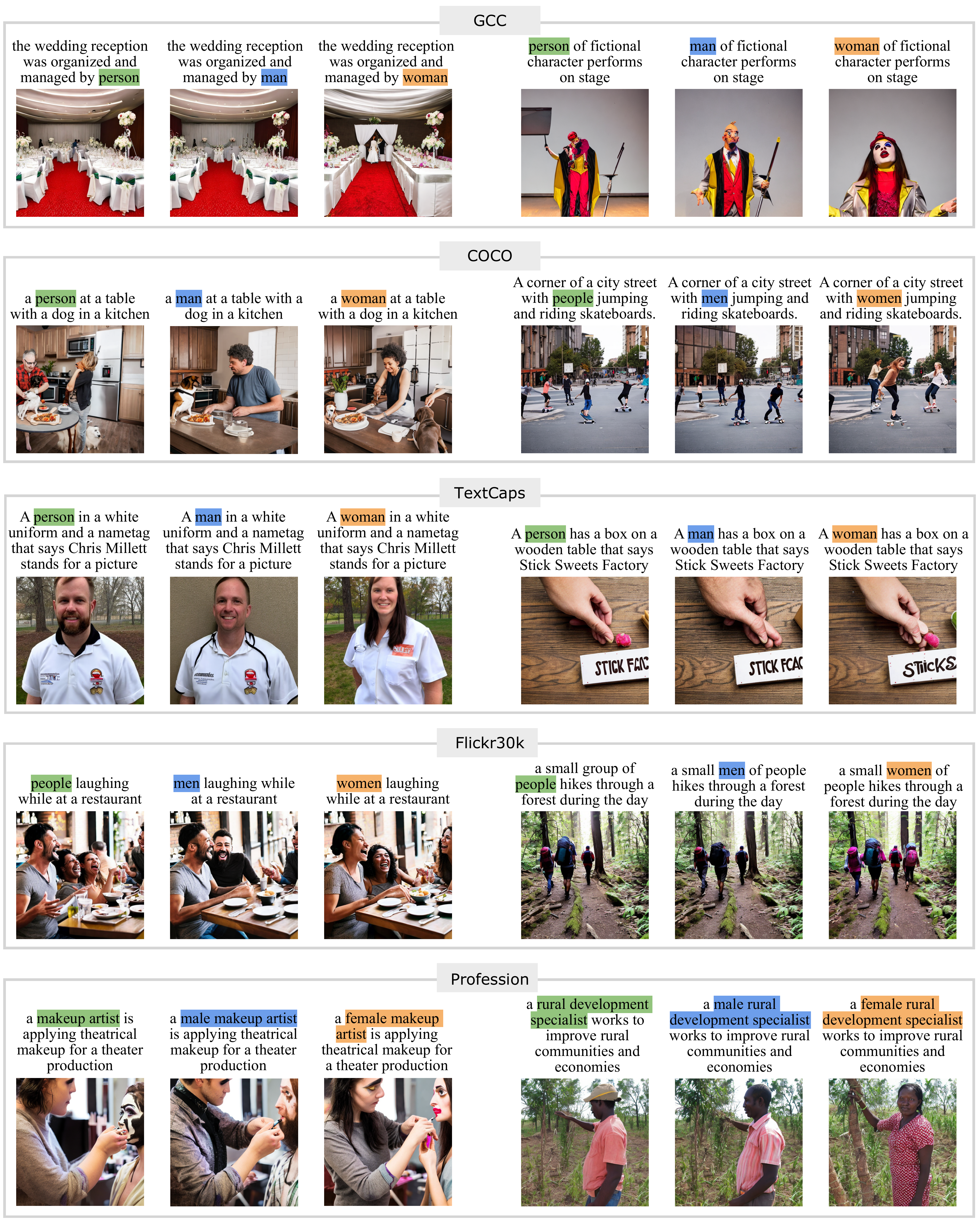}
\caption{Examples of triplet prompts and the corresponding generated images for each dataset on SD v2.0.}
\label{fig:supp_samples}
\end{figure*}

\paragraph{Examples}
Figure \ref{fig:supp_samples} shows examples of triplet prompts and the corresponding generated images for each dataset.

\paragraph{Profession names}
Table \ref{tab:supp_occupation} presents lists of the profession names used for generating prompts in the Profession set.

\begin{table*}[t]
\renewcommand{\arraystretch}{1.3}
\setlength{\tabcolsep}{7pt}
\footnotesize
\vspace{-8pt}
\centering

\begin{tabularx}{1.03\textwidth}{@{}p{0.2\textwidth}  p{0.8\textwidth}}
\toprule
Topic & Profession name \\
\midrule

Science & Botanist, Geologist, Oceanographer, Astronomer, Meteorologist, Chemist, Physicist, Geneticist, Archaeologist, Biostatistician, Marine Biologist, Quantum Physicist, Seismologist, Ecologist, Geophysicist, Epidemiologist, Materials Scientist, Neuroscientist, Volcanologist, Zoologist \\
\hline
Art & Street Artist, Songwriter, Calligrapher, Art Appraiser, Tattoo Artist, Mural Artist, Writer, Illustrator, Film Director, Ceramic Artist, Curator, Makeup Artist, Graffiti Artist, Furniture Designer, Cartoonist, Sculptor, Fashion Designer, Glassblower, Landscape Painter, Storyboard Artist \\
\hline
Sports & Athlete, Gymnast, Swimmer, Runner, Cyclist, Skier, Diver, Wrestler, Boxer, Surfer, Coach, Fitness Instructor, Sports Photographer, Referee, Sports Agent, Soccer Player, Tennis Coach, Yoga Instructor, Martial Arts Instructor, Golf Caddy \\
\hline
Celebrations & Wedding Planner, Party Decorator, Event Caterer, Balloon Artist, Fireworks Technician, Event DJ, Wedding Officiant, Event Photographer, Costume Designer, Event Coordinator, Cake Decorator, Floral Designer, Lighting Technician, Ice Sculptor, Musician, Face Painter, Magician, Pyrotechnician, Caricature Artist, Audiovisual Technician \\
\hline
Education & School Principal, Librarian, Academic Advisor, Teaching Assistant, School Psychologist, Early Childhood Educator, Curriculum Developer, Educational Technologist, Special Education Teacher, School Counselor, Online Instructor, Music Teacher, Art Teacher, Mathematics Teacher, Science Teacher, History Teacher, Language Teacher, Physical Education Teacher, College Professor, Career Counselor \\
\hline
Healthcare & Nurse, Doctor, Therapist, Surgeon, Pharmacist, Midwife, Paramedic, Psychologist, Radiologist, Dentist, Orthopedic Surgeon, Oncologist, Pediatrician, Anesthesiologist, Dermatologist, Neurologist, Cardiologist, Chiropractor, Veterinarian, Respiratory Therapist \\
\hline
Technology & Data Analyst, Information Security Analyst, AI Ethics Researcher, Virtual Reality Developer, Quantum Computing Researcher, Ethical Hacker, Robotics Engineer, Software Developer, Database Administrator, Network Engineer, Machine Learning Engineer, Cybersecurity Consultant, Web Developer, Cloud Architect, Digital Marketing Specialist, IT Support Specialist, Game Developer, UI Designer, Biomedical Engineer, Tech Startup \\
\hline
Business and Finance & Business Analyst, Tax Consultant, Financial Planner, Corporate Risk Manager, Actuary, Import-Export Specialist, Accountant, Investment Analyst, Operations Manager, Management Trainer, Small Business Consultant, Financial Auditor, Financial Controller, Human Resources Manager, Marketing Manager, Real Estate Agent, Supply Chain Manager, Chief Financial Officer, Economist, Chief Executive Officer \\
\hline
Government and Public Service & Diplomatic Services Officer, Social Services Worker, Public Policy Analyst, Environmental Health Inspector, Fire Marshal, Immigration Officer, Park Ranger, Community Organizer, Census Bureau Statistician, Emergency Management Director, Social Worker, Police Officer, Public Health Inspector, Environmental Scientist, City Planner, Legislative Aide, Judge, Foreign Service Officer, Conservation Officer, Civil Servant \\
\hline
Agriculture and Farming & Organic Farming Consultant, Beekeeper, Nutritionist, Agricultural Inspector, Poultry Farmer, Soil Conservationist, Aquaculture Technician, Agricultural Economist, Irrigation Specialist, Farm Equipment Mechanic, Livestock Rancher, Horticulturist, Viticulturist, Dairy Farmer, Agricultural Researcher, Fishery Manager, Rural Development Specialist, Animal Breeder, Greenhouse Manager, Sustainable Agriculture Advocate \\
\hline
Environmental & Wildlife Biologist, Environmental Educator, Green Building Architect, Environmental Geologist, Air Quality Specialist, Water Quality Analyst, Forest Ranger, Marine Ecologist, Climate Change Analyst, Conservation Biologist, Park Naturalist, Wetland Scientist, Renewable Energy Specialist, Sustainability Consultant, Eco-Tourism Guide, Environmental Impact Analyst, Land Use Planner, Soil Scientist, Environmental Policy Analyst, Recycling Coordinator \\
\hline
Travel and Hospitality & Travel Agent, Tour Guide, Hotel Manager, Flight Attendant, Cruise Ship Staff, Concierge, Restaurant Manager, Sommelier, Travel Blogger, Amusement Park Entertainer, Culinary Tour Guide, Hotel Concierge, Resort Manager, Airport Operations Manager, Tourism Marketing Specialist, Hospitality Sales Manager, Bed and Breakfast Owner, Cabin Crew Member, Theme Park Performer, Hostel Manager \\
\hline
Media and Journalism & War Correspondent, Documentary Filmmaker, Social Media Influencer, Radio Show Host, Film Critic, Multimedia Journalist, Travel Photographer, Sports Anchor, News Producer, Investigative Journalist, Foreign Correspondent, Photojournalist, Columnist, Podcast Host, Public Relations Specialist, Media Critic, Weather Forecaster, Press Secretary, News Editor, TV News Reporter \\
\hline
Law and Legal & Lawyer, Intellectual Property Attorney, Criminal Psychologist, Legal Ethicist, Court Clerk, Arbitrator, Paralegal, Legal Secretary, Legal Consultant, Immigration Attorney, Family Law Mediator, Legal Aid Attorney, Bankruptcy Attorney, Legal Translator, Corporate Counsel, Tax Attorney, Civil Litigation Attorney, Legal Auditor, Criminal Defense Attorney, Judicial Law Clerk \\
\hline
Manufacturing and Industry & Quality Assurance Manager, Industrial Hygienist, Production Scheduler, CNC Machinist, Factory Inspector, Metallurgical Engineer, Assembly Line Worker, Process Improvement Specialist, Materials Handler, Manufacturing Engineer, Welder, Packaging Technician, Facilities Manager, Maintenance Technician, Logistics Coordinator, Lean Manufacturing Specialist, Safety Coordinator, Inventory Control Analyst, Machine Operator, Operations Supervisor \\
\hline
Culinary and Food Services & Food Safety Inspector, Mixologist, Chef, Brewery Master, Baker, Restaurant Critic, Sommelier, Food Scientist, Caterer, Nutritionist, Butcher, Pastry Chef, Culinary Instructor, Wine Taster, Gourmet Food Store Owner, Food Stylist, Coffee Roaster, Line Cook, Chocolatier, Food Truck Owner \\

\bottomrule
\end{tabularx}
\caption{Profession names in the Profession set.}
\label{tab:supp_occupation}
\end{table*}

\subsection{DAAM}
Here we provide details for computing the attention map of words by DAAM \cite{tang2022daam}. Let $\mathbf{P}$ be a matrix whose column $n$ is the word embedding corresponding to the word $n$ in $p$, and $H(\mathbf{z}_t)$ be a feature map of a certain block of Stable Diffusion's UNet for latent embedding $\mathbf{z}_t$ in the $t$-th denoising step. Cross-attention between $\mathbf{P}$ and $H(\mathbf{z}_t)$ is given by:
\begin{align}
    \mathbf{A}_t = \text{softmax}\left(\frac{\mathbf{Q} \mathbf{K}^\top}{\sqrt{d}}\right),
\end{align}
where $\mathbf{Q}$ and $\mathbf{K}$ are the query and key matrices given using linear layers $\mathbf{W}_\text{Q}$ and $\mathbf{W}_\text{K}$ as $\mathbf{Q} = \mathbf{W}_\text{Q} H(\mathbf{z}_t)$ and $\mathbf{K} = \mathbf{W}_\text{K} \mathbf{P}$ whose output dimensionality is $d$.\footnote{The index $t$ for denoising step is omitted for simplicity.} The heart of DAAM is $\mathbf{A}_t$, of which column $n$ is the attention map from word $n$ to each spatial position of feature map $H(\mathbf{z}_t)$. We aggregate the attention maps over UNet blocks, multiple attention heads, and denoising steps.

\section{Research question 1}
\label{sec:sup:rq1}

\begin{table*}[t]
\renewcommand{\arraystretch}{1.2}
\setlength{\tabcolsep}{6pt}
\small
\centering

\begin{tabularx}{0.95\textwidth}{@{}p{1mm} l  r c r c r r r r r c@{}}
\toprule

& & Prompt & & Denoising & & \multicolumn{6}{c}{Image}\\ 
 \cline{3-3}
 \cline{5-5}
 \cline{7-12}

\multicolumn{2}{l}{Pairs} & $\mathbf{t}$ & & $\mathbf{z}_0$ & & \textit{SSIM} $\uparrow$ & \textit{Diff.~Pix.}$\downarrow$ & \textit{ResNet} $\uparrow$ & \textit{CLIP} $\uparrow$ & \textit{DINO} $\uparrow$ & \textit{split-product} $\uparrow$ \\

\midrule
\rowcolor{oursrow}
\multicolumn{12}{l}{SD v1.4} \\

\multicolumn{12}{l}{COCO} \\
& (neutral, feminine)  & $0.920$  & & $0.778$  & & $0.568$ & $38.558$ & $0.866$ & $\textbf{0.8584}$ & $0.564$ & $0.957$ \\

& (neutral, masculine)  & $\textbf{0.942}$  & & $\textbf{0.796}$  & & $\textbf{0.592}$ & $\textbf{35.671}$ & $\textbf{0.873}$ & $0.8580$ & $\textbf{0.591}$ & $\textbf{0.959}$ \\

\multicolumn{12}{l}{TextCaps} \\
& (neutral, feminine)  & $0.931$  & & $0.747$  & & $0.461$ & $46.873$ & $0.853$ & $0.773$ & $0.530$ & $0.952$ \\

& (neutral, masculine)  & $\textbf{0.948}$ & &  $\textbf{0.768}$ &  & $\textbf{0.487}$ & $\textbf{43.599}$ & $\textbf{0.862}$ & $\textbf{0.786}$ & $\textbf{0.555}$ & $\textbf{0.954}$ \\

\multicolumn{12}{l}{Flickr30k} \\
& (neutral, feminine)  & $0.913$ & & $0.792$ & & $0.492$ & $44.010$ & $0.858$ & $\textbf{0.830}$ & $0.563$ & $0.959$ \\

& (neutral, masculine)  & $\textbf{0.931}$ & & $\textbf{0.804}$ & & $\textbf{0.518}$ & $\textbf{41.105}$ & $\textbf{0.865}$ & $0.828$ & $\textbf{0.587}$ & $\textbf{0.960}$ \\

\multicolumn{12}{l}{Profession} \\
& (neutral, feminine)  & $0.854$ & & $0.765$ & & $0.487$ & $45.006$ & $0.831$ & $0.830$ & $0.528$ & $0.948$ \\

& (neutral, masculine)  & $\textbf{0.862}$ & & $\textbf{0.783}$ & & $\textbf{0.508}$ & $\textbf{42.528}$ & $\textbf{0.843}$ & $\textbf{0.846}$ & $\textbf{0.555}$ & $\textbf{0.952}$ \\

\rowcolor{oursrow}
\multicolumn{12}{l}{SD v2.0} \\

\multicolumn{12}{l}{COCO} \\
& (neutral, feminine)  & $0.984$ & & $0.793$ & & $0.603$ & $34.10$ & $0.881$ & $\textbf{0.861}$ & $0.595$ & $0.9645$   \\

& (neutral, masculine)  & $\textbf{0.985}$ & & $\textbf{0.805}$ & & $\textbf{0.616}$ & $\textbf{32.50}$ & $\textbf{0.887}$ & $0.859$ & $\textbf{0.609}$ & $\textbf{0.9647}$  \\

\multicolumn{12}{l}{TextCaps} \\
& (neutral, feminine)  & $0.9846$ & & $0.745$ & & $0.502$ & $41.41$ & $0.861$ & $0.771$ & $0.536$ & $0.958$  \\

& (neutral, masculine)  & $\textbf{0.9854}$ & & $\textbf{0.767}$ & & $\textbf{0.530}$ & $\textbf{37.41}$ & $\textbf{0.874}$ & $\textbf{0.791}$ & $\textbf{0.570}$ & $\textbf{0.962}$   \\

\multicolumn{12}{l}{Flickr30k} \\
& (neutral, feminine)  & $0.982$ & & $0.801$ & & $0.541$ & $38.42$ & $0.871$ & $\textbf{0.833}$ & $0.584$  & $0.9685$  \\

& (neutral, masculine)  & $\textbf{0.983}$ & & $\textbf{0.809}$ & & $\textbf{0.559}$ & $\textbf{36.02}$ & $\textbf{0.874}$ & $0.826$ & $\textbf{0.601}$ & $\textbf{0.9686}$  \\

\multicolumn{12}{l}{Profession} \\
& (neutral, feminine)  & $\textbf{0.85784}$ & & $0.766$ & & $0.511$ & $42.41$ & $0.839$ & $0.846$ & $0.537$  & $0.952$ \\

& (neutral, masculine)  & $0.85783$ & & $\textbf{0.779}$ & & $\textbf{0.528}$ & $\textbf{40.71}$ & $\textbf{0.848}$ & $\textbf{0.857}$ & $\textbf{0.556}$ & $\textbf{0.953}$  \\

\rowcolor{oursrow}
\multicolumn{12}{l}{SD v2.1} \\

\multicolumn{12}{l}{COCO} \\
& (neutral, feminine)  & $0.984$ & & $0.763$ & & $0.569$ & $37.796$ & $0.8670$ & $\textbf{0.858}$ & $0.556$ & $0.955$ \\

& (neutral, masculine)  & $\textbf{0.985}$ & & $\textbf{0.780}$ & &  $\textbf{0.586}$ & $\textbf{35.632}$ & $\textbf{0.8747}$ & $0.853$ & $\textbf{0.575}$ & $\textbf{0.957}$ \\

\multicolumn{12}{l}{TextCaps} \\
& (neutral, feminine)  & $0.9846$ & & $0.713$ & & $0.456$ & $46.600$ & $0.838$ & $0.752$ & $0.492$ & $0.948$ \\

& (neutral, masculine)  & $\textbf{0.9854}$ & & $\textbf{0.747}$ & & $\textbf{0.483}$ & $\textbf{43.362}$ & $\textbf{0.851}$ & $\textbf{0.773}$ & $\textbf{0.524}$ & $\textbf{0.953}$ \\

\multicolumn{12}{l}{Flickr30k} \\
& (neutral, feminine)  & $0.982$ & & $0.772$ & & $0.499$ & $42.722$ & $0.853$ & $\textbf{0.823}$ & $0.544$ & $\textbf{0.9572}$ \\

& (neutral, masculine)  & $\textbf{0.983}$ & & $\textbf{0.784}$ & & $\textbf{0.511}$ & $\textbf{40.988}$ & $\textbf{0.857}$ & $0.813$ & $\textbf{0.555}$ & $0.9570$ \\

\multicolumn{12}{l}{Profession} \\
& (neutral, feminine)  & $\textbf{0.85784}$ & & $0.759$ & & $0.497$ & $44.173$ & $0.835$ & $0.856$ & $0.521$ & $0.945$ \\

& (neutral, masculine)  & $0.85783$ & & $\textbf{0.778}$ & & $\textbf{0.517}$ & $\textbf{41.796}$ & $\textbf{0.848}$ & $\textbf{0.87}$ & $\textbf{0.548}$ & $\textbf{0.947}$ \\

\bottomrule
\end{tabularx}
\caption{Representational disparities between neutral, feminine, and masculine prompts in the three spaces on Stable Diffusion models.}
\label{tab:supp_rq1_repre_dispar}
\end{table*}

Table \ref{tab:supp_rq1_repre_dispar} presents representational disparities in COCO, TextCaps, Flickr30k, and Profession. The same trend can be observed across all datasets.

\section{Research question 2}
\label{sec:sup:rq2}

\begin{figure*}[t]
\hspace{-16pt}
\centering
\includegraphics[width=1\textwidth]{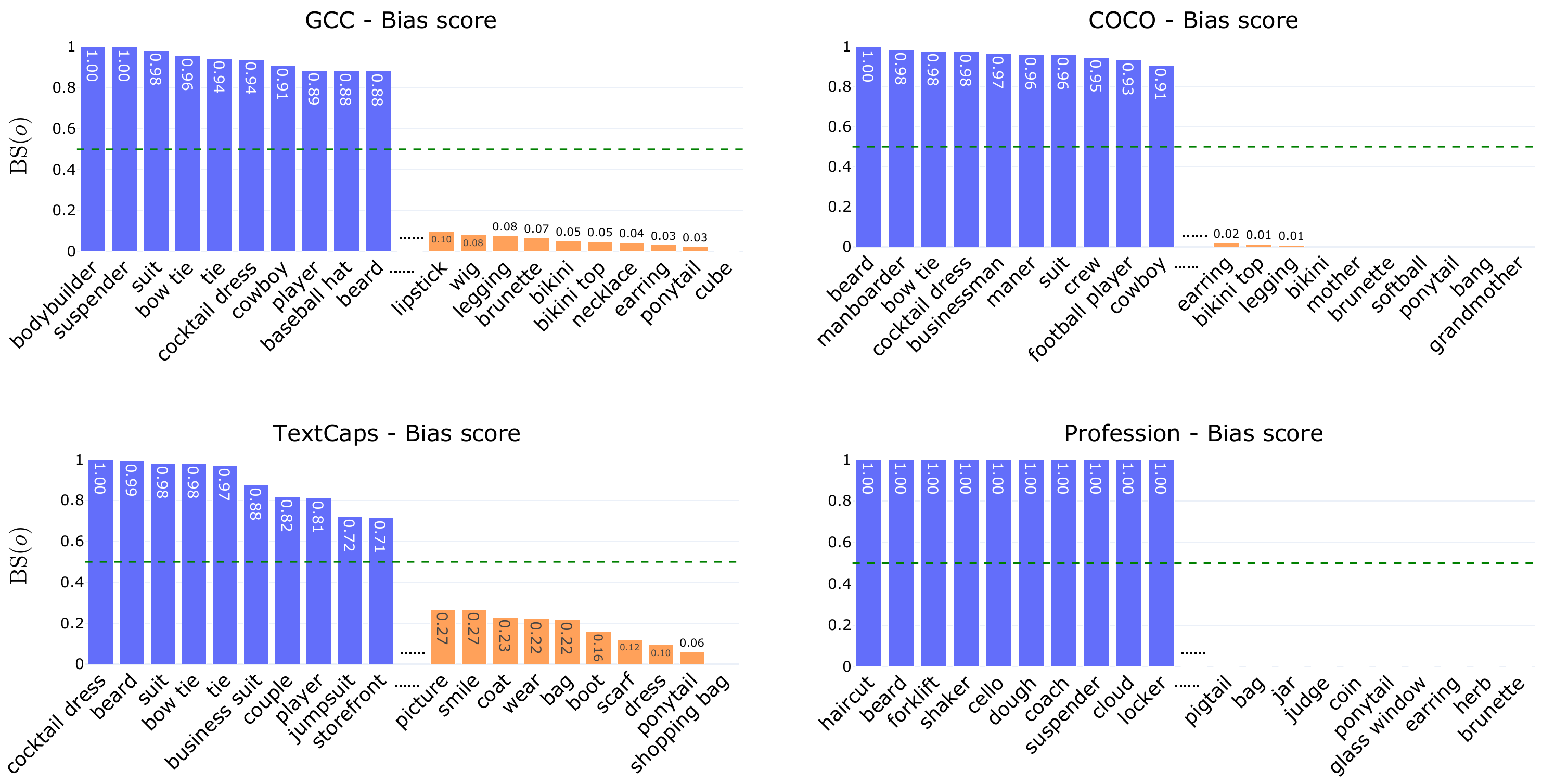}
\caption{Bias score in GCC, COCO, TextCaps, and Profession (SD v2.0).}
\label{fig:supp_rq2_bs_all}
\end{figure*}

\begin{figure*}[t]
\hspace{20pt}
\centering
\includegraphics[width=1\textwidth]{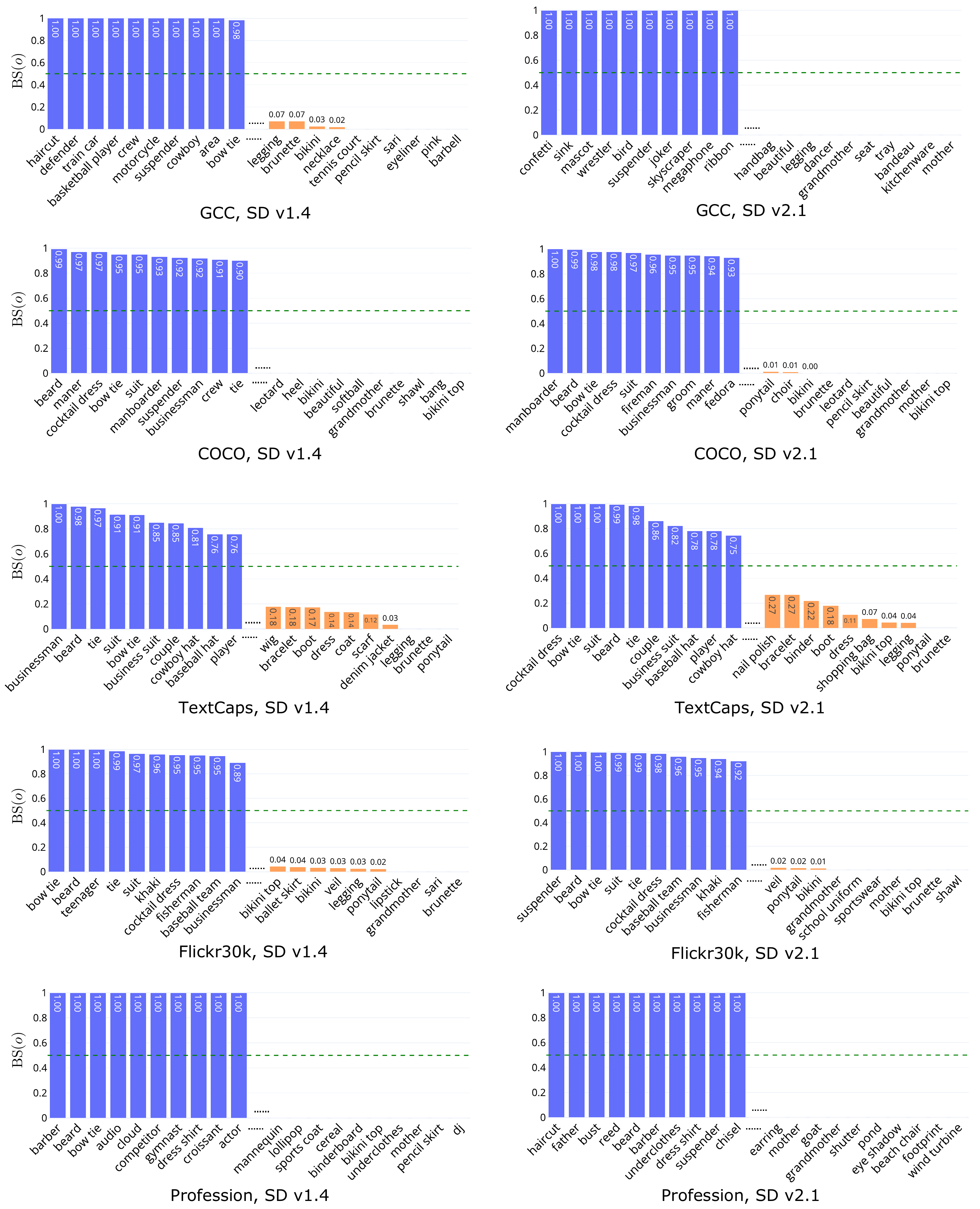}
\caption{Bias score on the datasets GCC, COCO, TextCaps, Flickr30k, and Profession, on SD v1.4 and SD v2.1.}
\label{fig:supp_v14n21_rq2_bs_all}
\end{figure*}

\paragraph{Bias score}
Figure \ref{fig:supp_rq2_bs_all} shows the bias score in other datasets on SD v2.0.
Figure \ref{fig:supp_v14n21_rq2_bs_all} shows the bias score on all datasets on SD v1.4 and SD v2.1.
These results exhibit a consistent trend across different datasets and models. For example, objects such as \texttt{beard}, \texttt{bow tie}, and \texttt{suit} consistently lean towards the \textit{masculine}, while \texttt{ponytail}, \texttt{brunette}, and \texttt{bikini} exhibit a preference for the \textit{feminine} in most datasets.

\section{Research question 3}
\label{sec:sup:rq3}
\begin{figure*}[t]
\hspace{-10pt}
\centering
\includegraphics[width=1\textwidth]{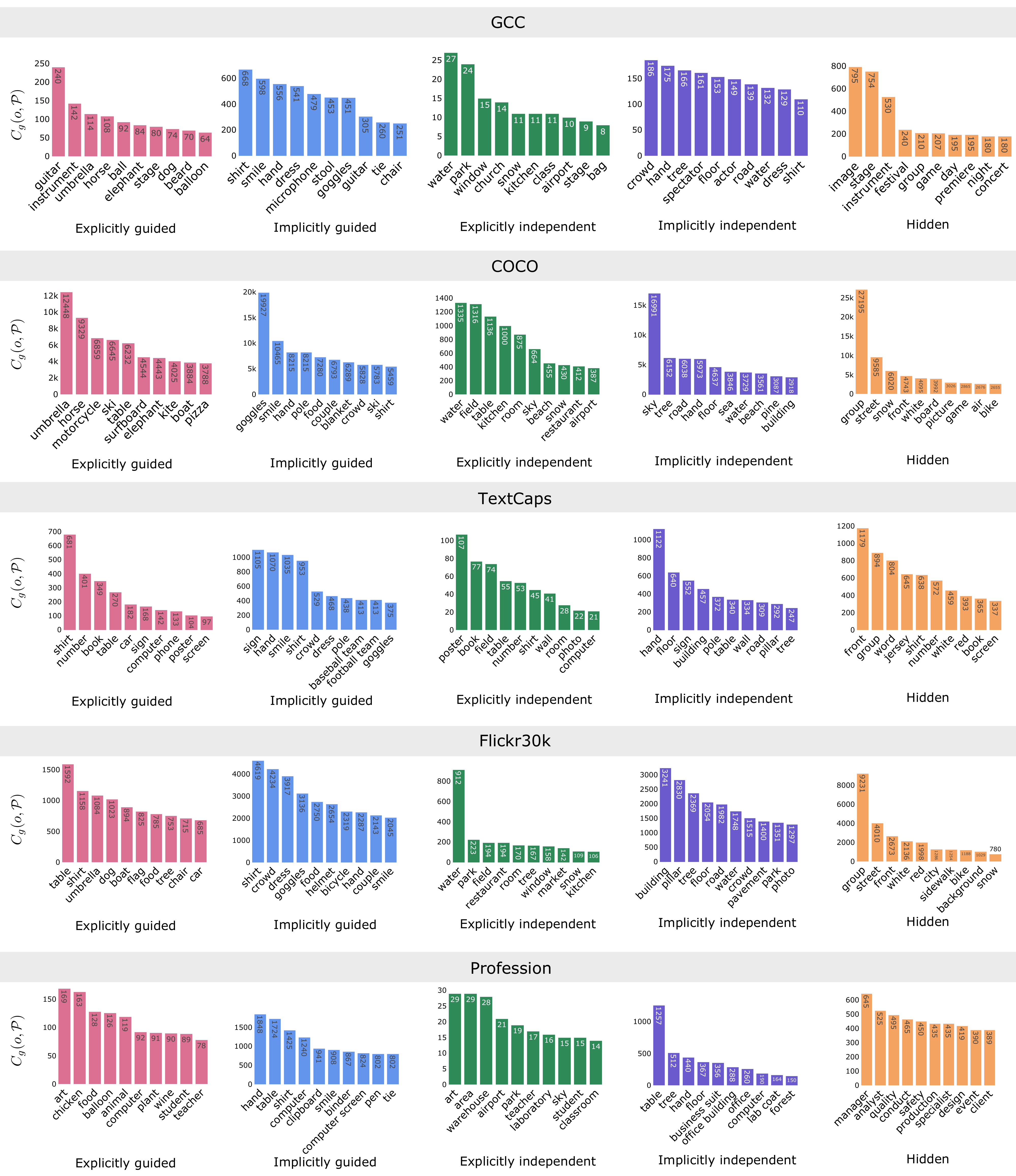}
\caption{The occurrence $C_g(o, \mathcal{P})$ of object $o$ in images generated from $\mathcal{P}$ on each dependency group for each dataset (SD v2.0).}
\label{fig:supp_rq3_cg_o}
\end{figure*}

\paragraph{Occurrence}
Figure \ref{fig:supp_rq3_cg_o} shows the occurrence $C_g(o, \mathcal{P})$ of the object $o$ in images generated from $\mathcal{P}$ on each prompt-image dependency group across all datasets.

\subsection{Chi-square test on dependency groups}
Table \ref{tab:supp_rq3_chi_group_v1.4}, \ref{tab:supp_rq3_chi_group_v2.0} and \ref{tab:supp_rq3_chi_group_v2.1} present chi-square test results on dependency groups in each dataset on SD v1.4, SD v2.0, and SD v2.1, respectively. It is observed that except for COCO, other datasets have a similar distribution in the \colorbox{g1}{explicitly guided} and \colorbox{g3}{explicitly independent} groups. Moreover, all datasets show significant differences in the \colorbox{g2}{implicitly guided} and \colorbox{g4}{implicitly independent} groups. Thus, we subsequently investigate objects in these two groups.

\begin{table*}[t]
\renewcommand{\arraystretch}{1.2}
\setlength{\tabcolsep}{7pt}
\small
\centering
\begin{tabularx}{0.72\textwidth}{@{} p{1mm} l r r r r r}
\toprule
& & Explicitly & Implicitly & Explicitly & Implicitly & \multirow{2}{*}{Hidden} \\
\multicolumn{2}{l}{SD v1.4} & guided & guided & independent & independent &  \\
\midrule

\rowcolor{oursrow}
\multicolumn{7}{l}{GCC} \\
& (neutral, feminine) & $0.761$ & \textbf{$< \num{e-5}$} & $0.293$ & \textbf{$< \num{e-5}$} & $1$ \\
& (neutral, masculine) & $0.607$ & \textbf{$< \num{e-5}$} & $0.805$ & \textbf{$< \num{e-5}$} & $1$ \\
& (feminine, masculine) & $0.865$ & \textbf{$< \num{e-5}$} & $0.605$ & \textbf{$< \num{e-5}$} & $1$ \\
& Triplet & $0.863$ & \textbf{$< \num{e-5}$} & $0.605$ & \textbf{$< \num{e-5}$} & $1$ \\

\rowcolor{oursrow}
\multicolumn{7}{l}{COCO} \\
& (neutral, feminine) & \textbf{$< \num{e-5}$} & \textbf{$< \num{e-5}$} & \textbf{$< \num{e-5}$} & \textbf{$< \num{e-5}$} & $1$ \\
& (neutral, masculine) & \textbf{$< \num{e-5}$} & \textbf{$< \num{e-5}$} & \textbf{$< \num{e-5}$} & \textbf{$< \num{e-5}$} & $1$ \\
& (feminine, masculine) & \textbf{$\num{4e-05}$} & \textbf{$< \num{e-5}$} & $\textbf{\num{2e-4}}$ & \textbf{$< \num{e-5}$} & $1$ \\
& Triplet & \textbf{$< \num{e-5}$} & \textbf{$< \num{e-5}$} & \textbf{$< \num{e-5}$} & \textbf{$< \num{e-5}$} & $1$ \\

\rowcolor{oursrow}
\multicolumn{7}{l}{TextCaps} \\
& (neutral, feminine) & $0.992$ & \textbf{$< \num{e-5}$} & $0.435$ & \textbf{$< \num{e-5}$} & $1$ \\
& (neutral, masculine) & $0.990$ & \textbf{$< \num{e-5}$} & $0.905$ & \textbf{$< \num{e-5}$} & $1$ \\
& (feminine, masculine) & $0.969$ & \textbf{$< \num{e-5}$} & $0.802$ & \textbf{$< \num{e-5}$} & $1$ \\
& Triplet & $0.999$ & \textbf{$< \num{e-5}$} & $0.778$ & \textbf{$< \num{e-5}$} & $1$ \\

\rowcolor{oursrow}
\multicolumn{7}{l}{Flickr30k} \\
& (neutral, feminine) & $0.654$ & \textbf{$< \num{e-5}$} & $0.297$ & \textbf{$< \num{e-5}$} & $1$ \\
& (neutral, masculine) & $0.858$ & \textbf{$< \num{e-5}$} & $0.330$ & \textbf{$< \num{e-5}$} & $1$ \\
& (feminine, masculine) & $0.858$ & \textbf{$< \num{e-5}$} & $0.650$ & \textbf{$< \num{e-5}$} & $1$ \\
& Triplet & $0.812$ & \textbf{$< \num{e-5}$} & $0.277$ & \textbf{$< \num{e-5}$} & $1$ \\

\rowcolor{oursrow}
\multicolumn{7}{l}{Profession} \\
& (neutral, feminine) & $0.598$ & \textbf{$< \num{e-5}$} & $0.288$ & \textbf{$< \num{e-5}$} & $1$ \\
& (neutral, masculine) & $0.431$ & \textbf{$< \num{e-5}$} & $0.755$ & \textbf{$< \num{e-5}$} & $1$ \\
& (feminine, masculine) & $0.380$ & \textbf{$< \num{e-5}$} & $0.497$ & \textbf{$< \num{e-5}$} & $1$ \\
& Triplet & $0.428$ & \textbf{$< \num{e-5}$} & $0.299$ & \textbf{$< \num{e-5}$} & $1$ \\

\bottomrule
\end{tabularx}
\caption{Chi-square test on dependency groups in each dataset on SD v1.4.}
\label{tab:supp_rq3_chi_group_v1.4}
\end{table*}

\begin{table*}[t]
\renewcommand{\arraystretch}{1.2}
\setlength{\tabcolsep}{7pt}
\small
\centering

\begin{tabularx}{0.77\textwidth}{@{} p{1mm} l r r r r r}
\toprule
& & Explicitly & Implicitly & Explicitly & Implicitly & \multirow{2}{*}{Hidden} \\
\multicolumn{2}{l}{SD v2.0} & guided & guided & independent & independent &  \\
\midrule

\rowcolor{oursrow}
\multicolumn{7}{l}{GCC} \\
& (neutral, feminine) & $0.196$ & \textbf{$< \num{e-5}$} & $0.191$ & \textbf{$< \num{e-5}$} & $1$ \\
& (neutral, masculine) & $0.878$ & \textbf{$< \num{e-5}$} & $0.690$ & $\textbf{\num{2e-3}}$ & $1$ \\
& (feminine, masculine) & $0.774$ & \textbf{$< \num{e-5}$} & $0.940$ & \textbf{$< \num{e-5}$} & $1$ \\
& Triplet & $0.751$ & \textbf{$< \num{e-5}$} & $0.653$ & \textbf{$< \num{e-5}$} & $1$ \\

\rowcolor{oursrow}
\multicolumn{7}{l}{COCO} \\
& (neutral, feminine) & \textbf{$< \num{e-5}$} & \textbf{$< \num{e-5}$} & \textbf{$< \num{e-5}$} & \textbf{$< \num{e-5}$} & $1$ \\
& (neutral, masculine) & \textbf{\num{1.01e-05}} & \textbf{$< \num{e-5}$} & \textbf{\num{3.05e-05}} & \textbf{$< \num{e-5}$} & $1$ \\
& (feminine, masculine) & \textbf{$< \num{e-5}$} & \textbf{$< \num{e-5}$} & $0.234$ & \textbf{$< \num{e-5}$} & $1$ \\
& Triplet & \textbf{$< \num{e-5}$} & \textbf{$< \num{e-5}$} & \textbf{$< \num{e-5}$} & \textbf{$< \num{e-5}$} & $1$ \\

\rowcolor{oursrow}
\multicolumn{7}{l}{TextCaps} \\
& (neutral, feminine) & $0.966$ & \textbf{$< \num{e-5}$} & $0.567$ & \textbf{$< \num{e-5}$} & $1$ \\
& (neutral, masculine) & $0.992$ & \textbf{$< \num{e-5}$} & $0.897$ & \textbf{\num{e-4}} & $1$ \\
& (feminine, masculine) & $0.990$ & \textbf{$< \num{e-5}$} & $0.551$ & \textbf{$< \num{e-5}$} & $1$ \\
& Triplet & $0.998$ & \textbf{$< \num{e-5}$} & $0.796$ & \textbf{$< \num{e-5}$} & $1$ \\

\rowcolor{oursrow}
\multicolumn{7}{l}{Flickr30k} \\
& (neutral, feminine) & $0.638$ & \textbf{$< \num{e-5}$} & $0.174$ & \textbf{$< \num{e-5}$} & $1$ \\
& (neutral, masculine) & $0.889$ & \textbf{$< \num{e-5}$} & $0.489$ & \textbf{$< \num{e-5}$} & $1$ \\
& (feminine, masculine) & $0.541$ & \textbf{$< \num{e-5}$} & $0.897$ & \textbf{$< \num{e-5}$} & $1$ \\
& Triplet & $0.704$ & \textbf{$< \num{e-5}$} & $0.391$ & \textbf{$< \num{e-5}$} & $1$ \\

\rowcolor{oursrow}
\multicolumn{7}{l}{Profession} \\
& (neutral, feminine) & $0.232$ & \textbf{$< \num{e-5}$} & $0.857$ & \textbf{$< \num{e-5}$} & $1$ \\
& (neutral, masculine) & $0.159$ & \textbf{$< \num{e-5}$} & $0.828$ & \textbf{$< \num{e-5}$} & $1$ \\
& (feminine, masculine) & $0.643$ & \textbf{$< \num{e-5}$} & $0.684$ & \textbf{$< \num{e-5}$} & $1$ \\
& Triplet & $0.235$ & \textbf{$< \num{e-5}$} & $0.929$ & \textbf{$< \num{e-5}$} & $1$ \\

\bottomrule
\end{tabularx}
\caption{Chi-square test on dependency groups in each dataset on SD v2.0.}
\label{tab:supp_rq3_chi_group_v2.0}
\end{table*}

\begin{table*}[t]
\renewcommand{\arraystretch}{1.2}
\setlength{\tabcolsep}{7pt}
\small
\centering

\begin{tabularx}{0.72\textwidth}{@{} p{1mm} l r r r r r}
\toprule
& & Explicitly & Implicitly & Explicitly & Implicitly & \multirow{2}{*}{Hidden} \\
\multicolumn{2}{l}{SD v2.1} & guided & guided & independent & independent &  \\
\midrule

\rowcolor{oursrow}
\multicolumn{7}{l}{GCC} \\
& (neutral, feminine) & $0.185$ & \textbf{$< \num{e-5}$} & $0.933$ & \textbf{$< \num{e-5}$} & $1$ \\
& (neutral, masculine) & $0.573$ & \textbf{$< \num{e-5}$} & $0.826$ & $\textbf{\num{3e-4}}$ & $1$ \\
& (feminine, masculine) & $0.942$ & \textbf{$< \num{e-5}$} & $0.714$ & \textbf{$< \num{e-5}$} & $1$ \\
& Triplet & $0.573$ & \textbf{$< \num{e-5}$} & $0.918$ & \textbf{$< \num{e-5}$} & $1$ \\

\rowcolor{oursrow}
\multicolumn{7}{l}{COCO} \\
& (neutral, feminine) & \textbf{$< \num{e-5}$} & \textbf{$< \num{e-5}$} & \textbf{$< \num{e-5}$} & \textbf{$< \num{e-5}$} & $1$ \\
& (neutral, masculine) & \textbf{$< \num{e-5}$} & \textbf{$< \num{e-5}$} & \textbf{$< \num{e-5}$} & \textbf{$< \num{e-5}$} & $1$ \\
& (feminine, masculine) & \textbf{$< \num{e-5}$} & \textbf{$< \num{e-5}$} & $0.056$ & \textbf{$< \num{e-5}$} & $1$ \\
& Triplet & \textbf{$< \num{e-5}$} & \textbf{$< \num{e-5}$} & \textbf{$< \num{e-5}$} & \textbf{$< \num{e-5}$} & $1$ \\

\rowcolor{oursrow}
\multicolumn{7}{l}{TextCaps} \\
& (neutral, feminine) & $0.839$ & \textbf{$< \num{e-5}$} & $0.234$ & \textbf{$< \num{e-5}$} & $1$ \\
& (neutral, masculine) & $0.994$ & \textbf{$< \num{e-5}$} & $0.467$ & \textbf{$< \num{e-5}$} & $1$ \\
& (feminine, masculine) & $0.972$ & \textbf{$< \num{e-5}$} & $0.941$ & \textbf{$< \num{e-5}$} & $1$ \\
& Triplet & $0.993$ & \textbf{$< \num{e-5}$} & $0.686$ & \textbf{$< \num{e-5}$} & $1$ \\

\rowcolor{oursrow}
\multicolumn{7}{l}{Flickr30k} \\
& (neutral, feminine) & $0.186$ & \textbf{$< \num{e-5}$} & $0.361$ & \textbf{$< \num{e-5}$} & $1$ \\
& (neutral, masculine) & $0.113$ & \textbf{$< \num{e-5}$} & $0.356$ & \textbf{$< \num{e-5}$} & $1$ \\
& (feminine, masculine) & $0.539$ & \textbf{$< \num{e-5}$} & $0.926$ & \textbf{$< \num{e-5}$} & $1$ \\
& Triplet & $0.109$ & \textbf{$< \num{e-5}$} & $0.504$ & \textbf{$< \num{e-5}$} & $1$ \\

\rowcolor{oursrow}
\multicolumn{7}{l}{Profession} \\
& (neutral, feminine) & $0.428$ & \textbf{$< \num{e-5}$} & $0.677$ & \textbf{$< \num{e-5}$} & $1$ \\
& (neutral, masculine) & $0.470$ & \textbf{$< \num{e-5}$} & $0.603$ & \textbf{$< \num{e-5}$} & $1$ \\
& (feminine, masculine) & $0.263$ & \textbf{$< \num{e-5}$} & $0.677$ & \textbf{$< \num{e-5}$} & $1$ \\
& Triplet & $0.338$ & \textbf{$< \num{e-5}$} & $0.703$ & \textbf{$< \num{e-5}$} & $1$ \\

\bottomrule
\end{tabularx}
\caption{Chi-square test on dependency groups in each dataset on SD v2.1.}
\label{tab:supp_rq3_chi_group_v2.1}
\end{table*}

\subsection{Bias score on implicitly guided}
\begin{figure*}[t]
\hspace{20pt}
\centering
\includegraphics[width=1\textwidth]{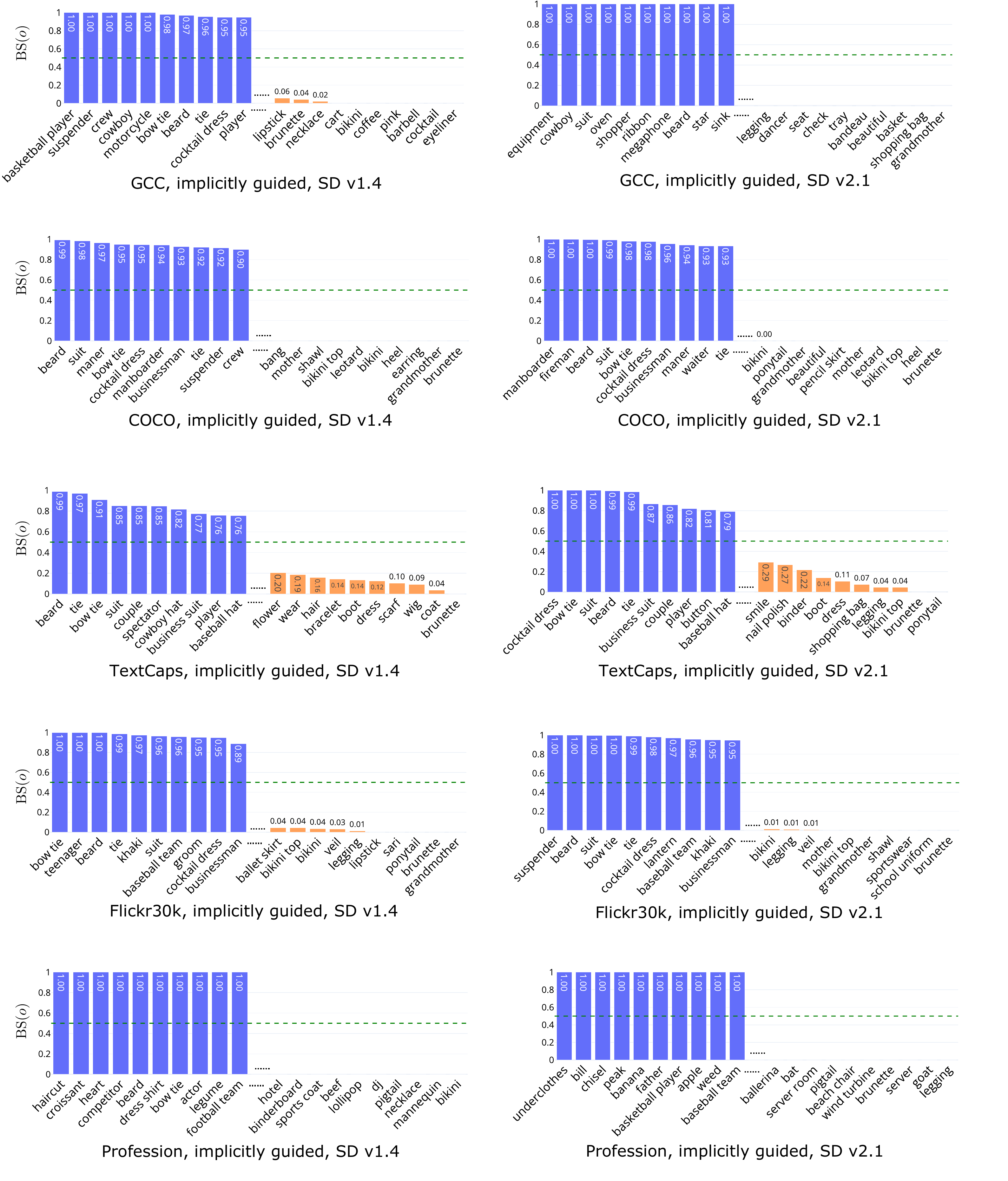}
\caption{Bias score on \colorbox{g2}{implicitly guided} in the datasets GCC, COCO, TextCaps, Flickr30k, and Profession, on SD v1.4 and SD v2.1.}
\label{fig:supp_v14n21_rq3_g2}
\end{figure*}

Figure \ref{fig:supp_rq3_g2} shows bias score of objects on \colorbox{g2}{implicitly guided} in the datasets GCC, COCO, Flickr30k, and Profession on SD v2.0. Figure \ref{fig:supp_v14n21_rq3_g2} shows bias score of objects on \colorbox{g2}{implicitly guided} on SD v1.4 and SD v2.1. We filter objects when the maximum co-occurrence is less than $5$ in GCC and Profession, and $20$ in COCO, TextCaps, and Flickr30k. The same setting is applied in the paper. It is observed that \texttt{suspender}, \texttt{beard}, and \texttt{bow tie} are more prone to appear in the \textit{masculine} when they are not mentioned in the prompt. Conversely, \texttt{ponytail}, \texttt{bikini}, and \texttt{leggings} are more associated with the \textit{feminine} than the \textit{masculine}.

\subsection{Bias score on implicitly independent}
\begin{figure*}[t]
    \centering
    \begin{subfigure}{\textwidth}
        \centering
        \includegraphics[width=1\textwidth]{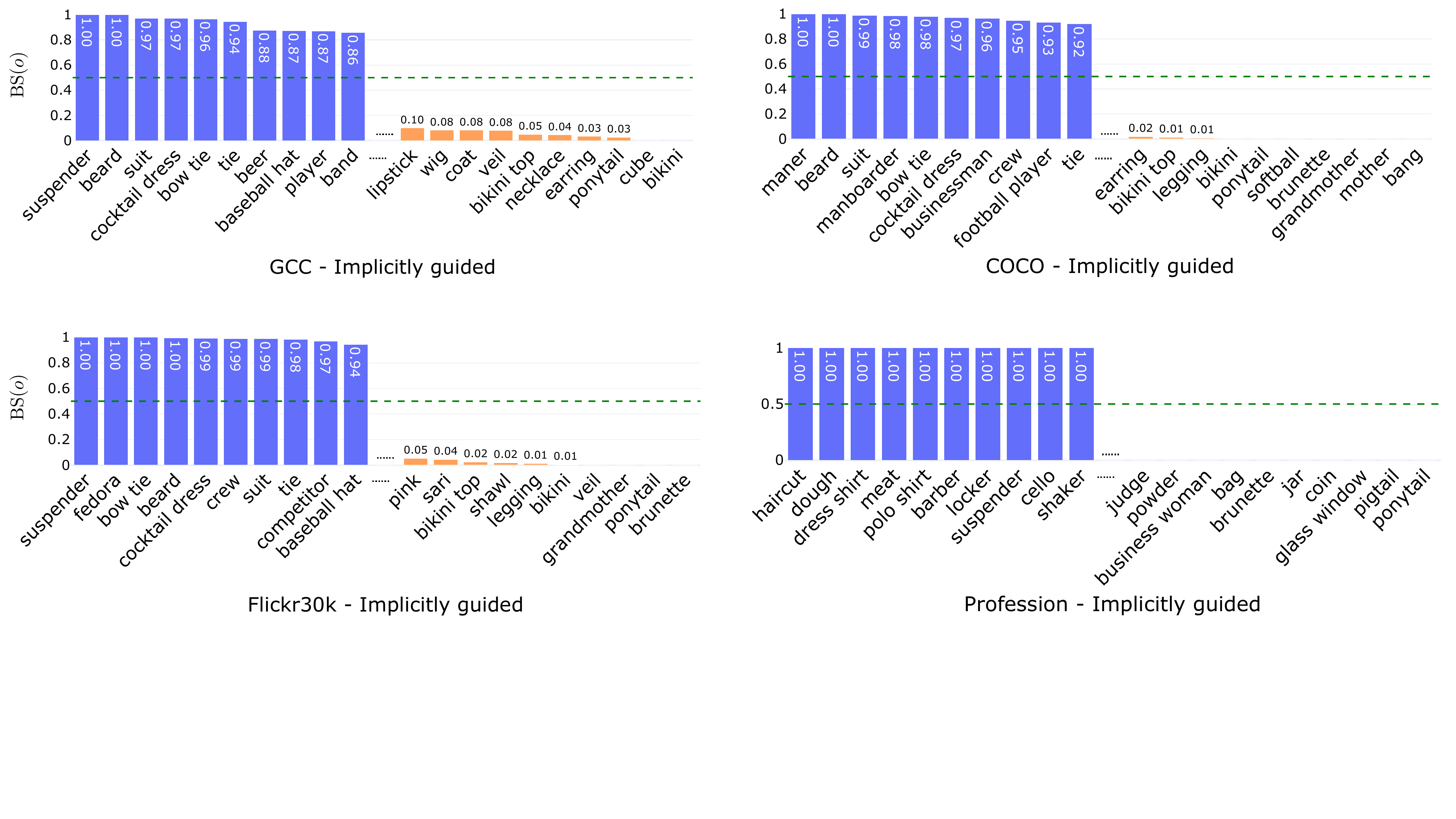}
        \caption{Bias score on \colorbox{g2}{implicitly guided}.}
        \label{fig:supp_rq3_g2}
    \end{subfigure}
    \begin{subfigure}{\textwidth}
        \centering
        \includegraphics[width=1\textwidth]{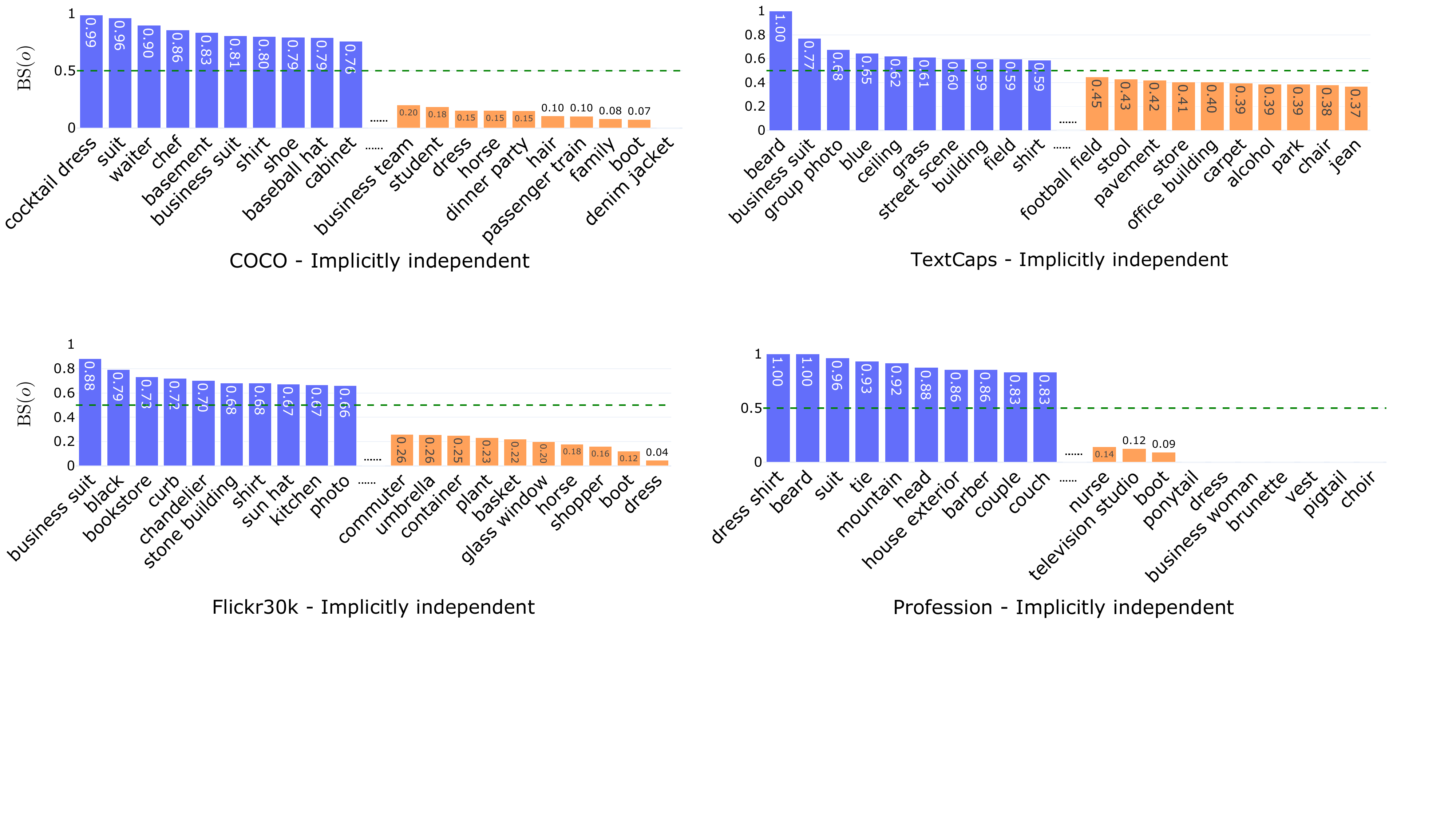}
        \caption{Bias score on \colorbox{g4}{implicitly independent}.}
        \label{fig:supp_rq3_g4}
    \end{subfigure} 
    \caption{Bias score on \colorbox{g2}{implicitly guided} and \colorbox{g4}{implicitly independent} on SD v2.0.}
\label{fig:supp_rq3_g2_g4}
\end{figure*}

\begin{figure*}[t]
\hspace{20pt}
\centering
\includegraphics[width=1\textwidth]{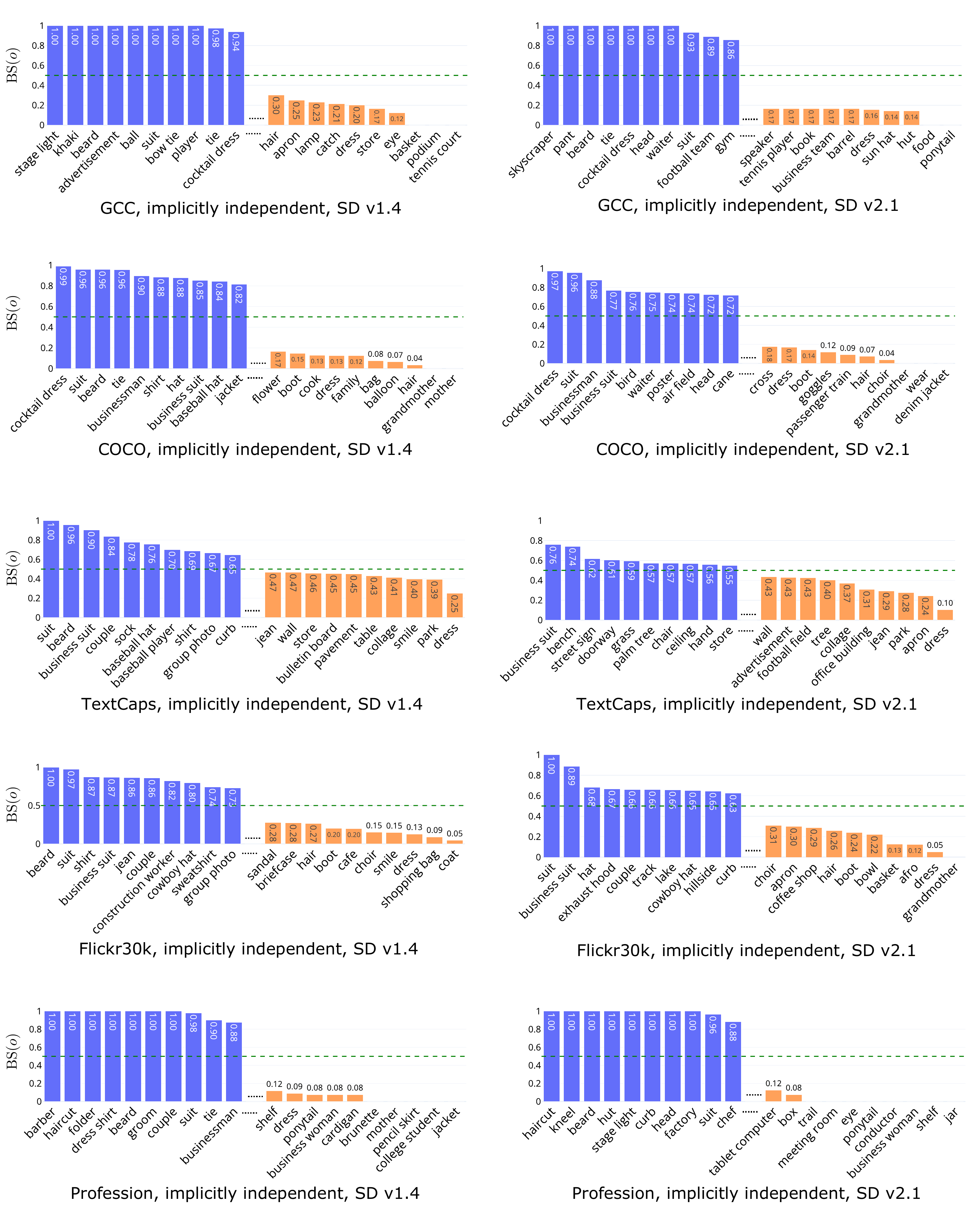}
\caption{Bias score on \colorbox{g4}{implicitly independent} in the datasets GCC, COCO, TextCaps, Flickr30k, and Profession, on SD v1.4 and SD v2.1.}
\label{fig:supp_v14n21_rq3_g4}
\end{figure*}

Figure \ref{fig:supp_rq3_g4} shows bias score of objects in \colorbox{g4}{implicitly independent} in the datasets COCO, TextCaps, Flickr30k, and Profession. Figure \ref{fig:supp_v14n21_rq3_g4} shows bias score of objects in \colorbox{g4}{implicitly independent} on SD v1.4 and SD v2.1.
Other than clothing, it shows that places and surroundings are most frequently associated with the \colorbox{g4}{implicitly independent} group. The specific environment is influenced by the semantics of the text. Therefore, we conduct analysis based on datasets. In COCO, results show that the \texttt{basement} and \texttt{cabinet} are more prone to appear in the \textit{masculine}, while \texttt{dinner party}, and \texttt{passenger train} are inclined to be generated in the \textit{feminine}. In TextCaps, \texttt{grass}, \texttt{building}, and \texttt{field} are skewed toward the \textit{masculine}, while \texttt{park}, \texttt{carpet}, and \texttt{store} are skewed toward the \textit{feminine}.

\section{Intra-prompt evaluation}
\label{sec:sup:intra}
Here we provide the results of the intra-prompt evaluation.
\begin{table*}[t]
\renewcommand{\arraystretch}{1.2}
\setlength{\tabcolsep}{6pt}
\small
\centering

\begin{tabularx}{0.91\textwidth}{@{}l  r c r c r r r r r c@{}}
\toprule

& Prompt & & Denoising & & \multicolumn{6}{c}{Image}\\ 
 \cline{2-2}
 \cline{4-4}
 \cline{6-11}

Pairs & $\mathbf{t}$ & & $\mathbf{z}_0$ & & \textit{SSIM} $\uparrow$ & \textit{Diff.~Pix.}$\downarrow$ & \textit{ResNet} $\uparrow$ & \textit{CLIP} $\uparrow$ & \textit{DINO} $\uparrow$ & \textit{split-product} $\uparrow$ \\

\midrule

(neutral, feminine)  & $0.981$ & & $0.789$ & & $0.547$ & $37.54$ & $0.867$ & $0.844$ & $0.557$ & $0.947$  \\

(neutral, masculine)  & $\textbf{0.982}$ & & $\textbf{0.829}$ & & $\textbf{0.587}$ & $\textbf{33.81}$ & $\textbf{0.892}$ & $\textbf{0.864}$ & $\textbf{0.625}$ & $\textbf{0.959}$  \\

\bottomrule
\end{tabularx}
\caption{Representational disparities between the neutral, feminine, and masculine in the three spaces from intra-prompts (SD v2.0).}
\label{tab:supp_intra_rq1}
\end{table*}

\paragraph{Representational Disparities}
The representational disparities in Table \ref{tab:supp_intra_rq1} show that the \textit{neutral} is consistently closer to \textit{masculine} in each space.

\section{Dependency groups analysis}
\label{sec:sup:groups_analysis}

\subsection{Dependency groups analysis}
Taking Stable Diffusion v2.0 as an example, we scrutinize the dependency groups deeper to discover the underlying connections between groups and objects.
\begin{table*}[t]
\renewcommand{\arraystretch}{1.1}
\setlength{\tabcolsep}{7pt}
\small
\centering

\begin{tabularx}{0.6\textwidth}{l  r  r  r  r  r}
\toprule
\multirow{2}{*}{Dataset} & Explicitly & Implicitly & Explicitly & Implicitly & \multirow{2}{*}{Hidden} \\
 & guided & guided & independent & independent &  \\
\midrule 

GCC & $64.48$ & $90.70$ & $7.81$ & $59.11$ & $96.14$ \\
COCO & $83.67$ & $93.54$ & $10.47$ & $57.53$ & $92.61$ \\
TextCaps & $61.97$ & $86.60$ & $8.78$ & $61.90$ & $99.10$ \\
Flickr30k & $83.07$ & $94.91$ & $9.56$ & $58.89$ & $92.48$ \\
Profession & $15.03$ & $98.07$ & $3.48$ & $63.22$ & $100.00$ \\

\bottomrule
\end{tabularx}
\caption{The proportion of images containing the dependency groups to all the images for each dataset on SD v2.0.}
\label{tab:supp_percent_image_in_groups}
\end{table*}

\paragraph{Image containing group}
To assess the presence of dependency groups in images, we compute the percentage of the images containing dependency groups over the total number of images in each dataset. The results are presented in Table \ref{tab:supp_percent_image_in_groups}. For example, in the GCC dataset, $64.48\%$ images contain at least one object in the \colorbox{g1}{explicitly guided} group. Similarly, in other datasets, over $60\%$ of images have objects in the \colorbox{g1}{explicitly guided}, except for the Profession set, where the proportion is only $15\%$. This disparity may be due to the specialized terminology in the Profession set, potentially reducing the chance of being detected by the visual grounding model. 
Conversely, only around $10\%$ or fewer images include objects in the \colorbox{g3}{explicitly independent} group. Given that most objects in this group represent the surrounding environment (as discussed in Sec. \ref{sec:analysis}), objects in \colorbox{g3}{explicitly independent} may occur when the prompt contains words indicating the surrounding environment (e.g., \texttt{park}, \texttt{kitchen}).

Moreover, a similar trend is observed across all datasets, where most images contain objects from the \colorbox{g2}{implicitly guided}, \colorbox{g4}{implicitly independent}, and \colorbox{g5}{hidden} group. This indicates that text-to-image models generate auxiliary objects to fill in both the areas guided by the prompt and those independent from it. 
We posit that the high proportion of \colorbox{g5}{hidden} group may be due to the abstract words that are challenging to detect and to the mismatch in synonyms. For instance, the visual grounding model may struggle to identify people as professions in the Profession set.

\paragraph{Amount of objects}
Next, we investigate the amount of individual objects in each dependency group and nouns in prompts. The results for each dataset are shown in Table \ref{tab:supp_num_obj_in_groups}. 
Supporting the findings in Table \ref{tab:supp_percent_image_in_groups}, objects in the \colorbox{g1}{explicitly guided} and \colorbox{g3}{explicitly independent} constitute only a small portion of the nouns in the prompts.
Besides, despite not being mentioned in the prompt, \colorbox{g2}{implicitly guided} and \colorbox{g4}{implicitly independent} groups contain more objects than \textit{explicitly} groups present in the image. This suggests that these two \textit{implicitly} groups are worth further exploration for a comprehensive understanding of the image generation process.
\begin{table*}[t]
\renewcommand{\arraystretch}{1.1}
\setlength{\tabcolsep}{7pt}
\small
\centering
\begin{tabularx}{0.68\textwidth}{l  r  r  r  r  r  r}
\toprule
\multirow{2}{*}{Dataset} & Explicitly & Implicitly & Explicitly & Implicitly & \multirow{2}{*}{Hidden} & \multirow{2}{*}{Nouns}  \\
 & guided & guided & independent & independent &  & \\
\midrule 

GCC & $155$ & $1,059$ & $85$ & $625$ & $536$ & $544$ \\
COCO & $827$ & $2,418$ & $391$ & $1,529$ & $3,274$ & $3,305$  \\
TextCaps & $371$ & $1,347$ & $147$ & $741$ & $3,608$ & $3,638$ \\
Flickr30k & $659$ & $2,017$ & $330$ & $1,255$ & $2,718$ & $2,741$  \\
Profession & $162$ & $1,331$ & $76$ & $650$ & $1,041$ & $1,043$  \\

\bottomrule
\end{tabularx}
\vspace{-5pt}
\caption{Amount of individual objects in each dependency group and nouns in prompts on SD v2.0 for each dataset.}
\label{tab:supp_num_obj_in_groups}
\vspace{-5pt}
\end{table*}

\paragraph{Group intersection ratio}
To uncover the common patterns and potential connections among the groups, we report the intersection ratio of individual objects among dependency groups and nouns in the prompts in Table \ref{tab:supp_intersection}. 
The ratio in each cell is computed from the intersection of two groups over the group in the stub column. For example, in GCC dataset, $76.77\%$ of objects in the \colorbox{g1}{explicitly guided} are also included in the \colorbox{g2}{implicitly guided} group.

Similar trends are observed across all datasets. Thus, we take GCC as an example.
We observe that many objects in the \textit{explicitly} groups are also found in the \textit{implicitly} groups. This suggests that these objects are more likely to be generated and detected, even if they are not explicitly mentioned in the prompts. Additionally, most objects in the \textit{explicitly} groups are also included in the \colorbox{g5}{hidden} group. This could be due to potential mismatches in synonyms, and there may be cases where the objects are not generated or detected.

On the other hand, within the \textit{implicitly} groups, only a small fraction (about 10\%) of objects are also present in the \textit{explicitly} group. This indicates that these objects are used to fill the scene but are not likely to be explicitly mentioned in the prompt. For example, with a prompt like ``\texttt{a person is walking along the street}'', it is common if the generated image contains pavement. However, \texttt{pavement} might not be explicitly mentioned unless the scene specifically relates to it, such as ``\texttt{a person is crossing the pavement}''.

\begin{table*}[t]
\renewcommand{\arraystretch}{1.2}
\setlength{\tabcolsep}{7pt}
\small
\centering

\begin{tabularx}{0.83\textwidth}{@{} p{1mm} l r r r r r r}
\toprule

 & & Explicitly & Implicitly & Explicitly & Implicitly & \multirow{2}{*}{Hidden} & \multirow{2}{*}{Nouns} \\
 & & guided & guided & independent & independent & & \\
\midrule

\rowcolor{oursrow}
\multicolumn{8}{l}{GCC} \\
& Over Explicitly guided & \cellcolor{darker}$100.00$ & \cellcolor{lighter}$76.77$ & $48.39$ & \cellcolor{lighter}$66.45$ & \cellcolor{darker}$94.84$ & \cellcolor{darker}$100.00$ \\
& Over Implicitly guided & $11.24$ & \cellcolor{darker}$100.00$ & $0.63$ & $48.35$ & $14.16$ & $14.83$ \\
& Over Explicitly independent & \cellcolor{darker}$88.24$ & \cellcolor{lighter}$78.82$ & \cellcolor{darker}$100.00$ & \cellcolor{lighter}$75.29$ & \cellcolor{darker}$92.94$ & \cellcolor{darker}$100.00$ \\
& Over Implicitly independent & $16.48$ & \cellcolor{darker}$81.92$ & $10.24$ & \cellcolor{darker}$100.00$ & $22.08$ & $23.04$ \\
& Over Hidden & $27.43$ & $27.99$ & $14.74$ & $25.75$ & \cellcolor{darker}$100.00$ & \cellcolor{darker}$100.00$ \\
& Over Nouns & $28.49$ & $28.86$ & $15.62$ & $26.47$ & \cellcolor{darker}$98.53$ & \cellcolor{darker}$100.00$ \\

\rowcolor{oursrow}
\multicolumn{8}{l}{COCO} \\
& Over Explicitly guided & \cellcolor{darker}$100.00$ & \cellcolor{darker}$93.95$ & $44.98$ & \cellcolor{lighter}$78.72$ & \cellcolor{darker}$96.37$ & \cellcolor{darker}$100.00$ \\
& Over Implicitly guided & $32.13$ & \cellcolor{darker}$100.00$ & $15.67$ & $58.35$ & $45.16$ & $46.36$ \\
& Over Explicitly independent & \cellcolor{darker}$95.14$ & \cellcolor{darker}$96.93$ & \cellcolor{darker}$100.00$ & \cellcolor{darker}$93.61$ & \cellcolor{darker}$99.23$ & \cellcolor{darker}$100.00$ \\
& Over Implicitly independent & $42.58$ & \cellcolor{darker}$92.28$ & $23.94$ & \cellcolor{darker}$100.00$ & $52.71$ & $54.35$ \\
& Over Hidden & $24.34$ & $33.35$ & $11.85$ & $24.62$ & \cellcolor{darker}$100.00$ & \cellcolor{darker}$100.00$ \\
& Over Nouns & $25.02$ & $33.92$ & $11.83$ & $25.14$ & \cellcolor{darker}$99.06$ & \cellcolor{darker}$100.00$ \\

\rowcolor{oursrow}
\multicolumn{8}{l}{TextCaps} \\
& Over Explicitly guided & \cellcolor{darker}$100.00$ & \cellcolor{darker}$86.79$ & $32.88$ & \cellcolor{lighter}$60.92$ & \cellcolor{darker}$91.91$ & \cellcolor{darker}$100.00$ \\
& Over Implicitly guided & $23.90$ & \cellcolor{darker}$100.00$ & $9.58$ & $47.29$ & $37.27$ & $39.20$ \\
& Over Explicitly independent & \cellcolor{darker}$82.99$ & \cellcolor{darker}$87.76$ & \cellcolor{darker}$100.00$ & \cellcolor{lighter}$76.87$ & \cellcolor{darker}$95.24$ & \cellcolor{darker}$100.00$ \\
& Over Implicitly independent & $30.50$ & \cellcolor{darker}$85.96$ & $15.25$ & \cellcolor{darker}$100.00$ & $44.13$ & $46.69$ \\
& Over Hidden & $9.50$ & $13.90$ & $3.88$ & $9.06$ & \cellcolor{darker}$100.00$ & \cellcolor{darker}$100.00$ \\
& Over Nouns & $10.20$ & $14.51$ & $4.04$ & $9.51$ & \cellcolor{darker}$99.18$ & \cellcolor{darker}$100.00$ \\

\rowcolor{oursrow}
\multicolumn{8}{l}{Flickr30k} \\
& Over Explicitly guided & \cellcolor{darker}$100.00$ & \cellcolor{darker}$92.56$ & $44.76$ & \cellcolor{lighter}$73.90$ & \cellcolor{darker}$96.81$ & \cellcolor{darker}$100.00$ \\
& Over Implicitly guided & $30.24$ & \cellcolor{darker}$100.00$ & $15.62$ & $55.97$ & $43.88$ & $44.72$ \\
& Over Explicitly independent & \cellcolor{darker}$89.39$ & \cellcolor{darker}$95.45$ & \cellcolor{darker}$100.00$ & \cellcolor{darker}$86.97$ & \cellcolor{darker}$97.27$ & \cellcolor{darker}$100.00$ \\
& Over Implicitly independent & $38.80$ & \cellcolor{darker}$89.96$ & $22.87$ & \cellcolor{darker}$100.00$ & $51.16$ & $52.11$ \\
& Over Hidden & $23.47$ & $32.56$ & $11.81$ & $23.62$ & \cellcolor{darker}$100.00$ & \cellcolor{darker}$100.00$ \\
& Over Nouns & $24.04$ & $32.91$ & $12.04$ & $23.86$ & \cellcolor{darker}$99.16$ & \cellcolor{darker}$100.00$ \\

\rowcolor{oursrow}
\multicolumn{8}{l}{Profession} \\
& Over Explicitly guided & \cellcolor{darker}$100.00$ & \cellcolor{darker}$81.48$ & $38.89$ & \cellcolor{lighter}$60.49$ & \cellcolor{darker}$98.77$ & \cellcolor{darker}$100.00$ \\
& Over Implicitly guided & $9.92$ & \cellcolor{darker}$100.00$ & $4.73$ & $42.15$ & $14.12$ & $14.27$ \\
& Over Explicitly independent & \cellcolor{darker}$82.89$ & \cellcolor{darker}$82.89$ & \cellcolor{darker}$100.00$ & \cellcolor{lighter}$75.00$ & \cellcolor{darker}$98.68$ & \cellcolor{darker}$100.00$ \\
& Over Implicitly independent & $15.08$ & \cellcolor{darker}$86.31$ & $8.77$ & \cellcolor{darker}$100.00$ & $20.31$ & $20.62$ \\
& Over Hidden & $15.37$ & $18.06$ & $7.20$ & $12.68$ & \cellcolor{darker}$100.00$ & \cellcolor{darker}$100.00$ \\
& Over Nouns & $15.53$ & $18.22$ & $7.29$ & $12.85$ & \cellcolor{darker}$99.81$ & \cellcolor{darker}$100.00$ \\
\bottomrule
\end{tabularx}
\caption{Intersection ratio of individual objects among dependency groups and nouns in the prompts on SD v2.0.}
\label{tab:supp_intersection}
\end{table*}

\clearpage

\end{document}